\documentclass[10pt,journal,compsoc]{IEEEtran}
\usepackage{amsmath,amsfonts}
\usepackage{algorithmic}
\usepackage{array}
\usepackage{textcomp}
\usepackage{stfloats}
\usepackage{url}
\usepackage{verbatim}
\usepackage{graphicx}
\usepackage[table]{xcolor}
\usepackage{epsfig}
\usepackage{graphicx}
\usepackage{amsmath}
\usepackage{amssymb}
\usepackage{enumitem}
\usepackage{pifont}

\usepackage{multirow}
\usepackage{tabularx}
\usepackage{tcolorbox}
\usepackage{dsfont}
\usepackage{url}
\usepackage[tight]{subfigure}
\usepackage{color}
\usepackage{subfigure}
\usepackage{amsthm}
\usepackage[export]{adjustbox}
\usepackage{comment}
\usepackage[utf8]{inputenc} 
\usepackage[T1]{fontenc}    
\usepackage{url}            
\usepackage{booktabs}       
\usepackage{amsfonts}       
\usepackage{nicefrac}       
\usepackage{microtype}      
\usepackage{diagbox}
\usepackage{balance}
\usepackage{subcaption}
\usepackage{sidecap}
\usepackage{tcolorbox}
\usepackage{pifont}
\usepackage{wrapfig}
\usepackage{longtable} 
\usepackage[linesnumbered,titlenumbered,ruled,vlined,commentsnumbered]{algorithm2e}
\SetKwComment{Comment}{$\triangleright$\ }{}
\definecolor{cvprblue}{rgb}{0.21,0.49,0.74}
\usepackage{siunitx}
\usepackage{mathtools}
\usepackage{soul}
\usepackage{stfloats}
\usepackage{colortbl}
\usepackage{bm}
\usepackage{makecell}
\usepackage{subcaption}
\usepackage{sidecap}
\usepackage{helvet}  
\usepackage{courier}  
\usepackage{babel}
\usepackage{array}

\usepackage[colorlinks,linkcolor=red,citecolor=blue]{hyperref}

\def\argmax{\operatornamewithlimits{arg\,max}}

\usepackage{xspace}
\makeatletter
\DeclareRobustCommand\onedot{\futurelet\@let@token\@onedot}
\def\@onedot{\ifx\@let@token.\else.\null\fi\xspace}
\def\eg{\emph{e.g}\onedot} 
\def\ie{\emph{i.e}\onedot}

\makeatother

\makeatletter
\renewcommand{\maketag@@@}[1]{\hbox{\m@th\normalsize\normalfont#1}}%
\makeatother

\ifCLASSOPTIONcompsoc
  \usepackage[nocompress]{cite}
\else
  \usepackage{cite}
\fi
\ifCLASSINFOpdf
\else
\fi
\begin{document}
%
\title{Scale-Invariant Adversarial Attack\\ against Arbitrary-scale Super-resolution}
%
%
%
%

\author{Yihao~Huang,
        Xin~Luo,
        Qing~Guo,
        Felix~Juefei-Xu,
        Xiaojun~Jia,
        Weikai~Miao,
        Geguang~Pu,
        and~Yang~Liu
}
\author{
    \IEEEauthorblockN{
        Yihao~Huang$^1$,
        Xin~Luo$^2$,
        Qing~Guo$^3$,
        Felix~Juefei-Xu$^4$,\\
        Xiaojun~Jia$^1$,
        Weikai~Miao$^2$,
        Geguang~Pu$^2$,
        and~Yang~Liu$^1$}\\
    \IEEEauthorblockA{$^1$ Nanyang Technological University, Singapore}\\
    \IEEEauthorblockA{$^2$ East China Normal University, China}\\
    \IEEEauthorblockA{$^3$ Agency for Science, Technology and Research (A*STAR), Singapore}\\
    \IEEEauthorblockA{$^4$ New York University, USA}\\
}

%
%

\markboth{}%
{Shell \MakeLowercase{\textit{et al.}}: Bare Advanced Demo of IEEEtran.cls for IEEE Computer Society Journals}
%



\IEEEtitleabstractindextext{%
\begin{abstract}
The advent of local continuous image function (LIIF) has garnered significant attention for arbitrary-scale super-resolution (SR) techniques. However, while the vulnerabilities of fixed-scale SR have been assessed, the robustness of continuous representation-based arbitrary-scale SR against adversarial attacks remains an area warranting further exploration. The elaborately designed adversarial attacks for fixed-scale SR are scale-dependent, which will cause time-consuming and memory-consuming problems when applied to arbitrary-scale SR. To address this concern, we propose a simple yet effective ``scale-invariant'' SR adversarial attack method with good transferability, termed \textbf{SIAGT}. Specifically, we propose to construct resource-saving attacks by exploiting finite discrete points of continuous representation. In addition, we formulate a coordinate-dependent loss to enhance the cross-model transferability of the attack. The attack can significantly deteriorate the SR images while introducing imperceptible distortion to the targeted low-resolution (LR) images. Experiments carried out on three popular LIIF-based SR approaches and four classical SR datasets show remarkable attack performance and transferability of SIAGT.
\end{abstract}

\begin{IEEEkeywords}
Adversarial Attack, Arbitrary-scale Super-resolution
\end{IEEEkeywords}}

\maketitle

\IEEEdisplaynontitleabstractindextext

%
\IEEEpeerreviewmaketitle

\ifCLASSOPTIONcompsoc
\IEEEraisesectionheading{\section{Introduction}\label{sec:introduction}}\label{sec:intro}
\else
\section{Introduction}
\label{sec:introduction}\label{sec:intro}
\fi

\IEEEPARstart{S}ingle image super-resolution (SISR) is a low-level computer vision task that aims to reconstruct a high-resolution image from a low-resolution (LR) one. A line of SISR research \cite{chen2021pre,dai2019second,liang2021swinir,mei2021image} referred to as ``fixed-scale'' SR focuses on extracting feature maps and leveraging the features to upsample images with a predefined scale. These methods need to train a series of models for each scale factor separately and constrain the potential applications when facing limited storage and computing resources. In contrast, learning continuous representations \cite{chen2021learning,xu2021ultrasr,liu2021enhancing,lee2022local,cao2023ciaosr} has recently gained popularity for constructing ``arbitrary-scale'' SR because of its ability to restore images in a continuous manner with only a single network.

For both fixed-scale or arbitrary-scale SR, assessing their vulnerability \cite{choi2019evaluating} is paramount since it helps for the development of more resilient and reliable SR models \cite{xun2024minimalism} that can perform effectively under a wide range of conditions \cite{castillo2021generalized,aakerberg2024pda}. In this paper, we analyze the SR models from the perspective of adversarial attack (usually exploited as a mainstream robustness evaluation method against DNN) \cite{goodfellow2014explaining,kurakin2016adversarial,madry2017towards,feng2023robust}. The robustness of fixed-scale SR models has been evaluated by previous work \cite{choi2019evaluating}. However, the method is a \textbf{scale-dependent} attack that is elaborately designed for fixed-scale SR tasks. It inevitably suffers from \textit{time-consuming} and \textit{memory-consuming} problems when applied to arbitrary-scale SR methods due to its special loss design based on fixed-scale SR outputs, leading to obvious practical shortcomings.

\begin{figure}[t]
\centering
\includegraphics[width=0.9\linewidth]{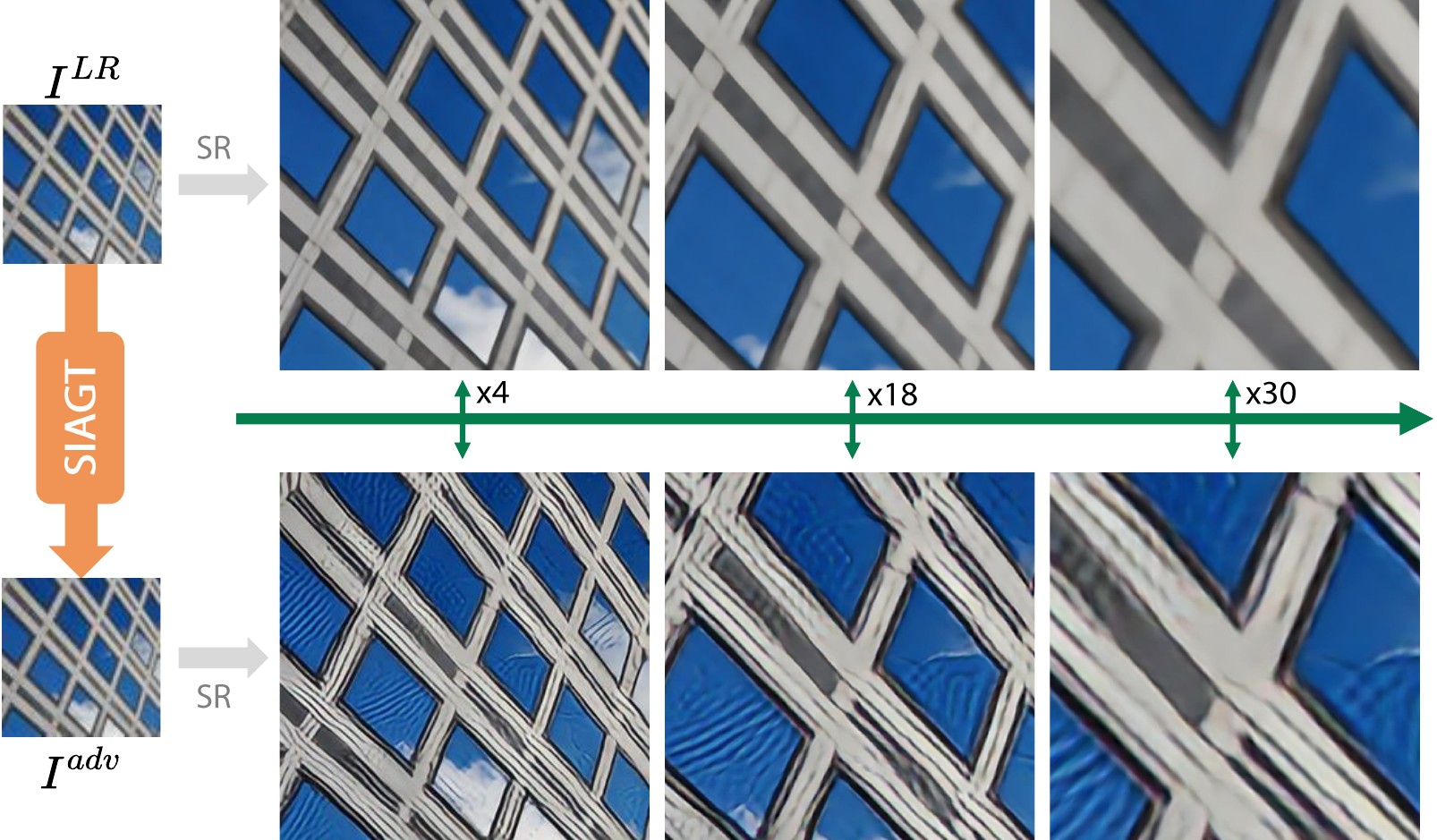}
\caption{A comparison of the SR output (by LIIF \cite{chen2021learning}) between the original LR image $I^{LR}$ and the adversarial image $I^{adv}$. Under arbitrary scales, the SR output of $I^{adv}$ generated by our attack method SIAGT are all deteriorated.}
\label{fig:introduction}
\end{figure}

To overcome the limitation and construct an appropriate adversarial attack for evaluating the robustness of arbitrary-scale SR methods, we propose a simple, resource-saving yet effective \textbf{Scale-Invariant} SR adversarial \textbf{A}ttack with \textbf{G}ood \textbf{T}ransferability, termed \textbf{SIAGT}. Please note that since almost all the arbitrary-scale SR approaches are based on continuous representation inspired by local continuous image function (LIIF) \cite{chen2021learning}, our research primarily designs adversarial attacks for continuous representation-based (or LIIF-based) SR methods. In order to make the attack inherently scale-invariant, instead of designing the loss based on \textit{strong scale-dependent SR outputs} as the previous method \cite{choi2019evaluating} did, the high-level idea of our method is to design the loss based on the \textit{scale-invariant continuous representation}. However, faced with a continuous image that can be \textit{infinitely} subdivided (\ie, has infinite coordinate points), traversing all coordinate points to build loss is impossible, which makes it a challenge to design the loss. To tackle this, we propose an attack method that first divides the continuous image into \textit{finite} blocks (\eg, the resolution of the LR image) and then samples \textit{finite} coordinate points per block for loss design. Through empirical study, we verify this method can \textit{efficiently} and \textit{effectively} attack arbitrary-scale SR methods for generating deteriorated SR output. Furthermore, cross-model transferability is an important property of white-box adversarial attacks and has not been investigated on SR tasks. Thus we further propose a coordinates-dependent loss to improve the transferability of our scale-invariant attack across continuous representation-based SR models. To sum up, our work has the following contributions:
\begin{itemize}
\item To our best knowledge, we are the first to propose a scale-invariant adversarial attack method for the robustness evaluation of the arbitrary-scale SR methods. The attack is simple yet effective and can generate adversarial examples under fixed time and memory consumption.
\item We propose a coordinate-dependent loss that can stably improve the cross-model transferability of the attack on continuous representation-based SR methods. 
\item The experiments and ablation study conducted on three LIIF-based SR methods and four classical SR datasets show the efficiency and effectiveness of SIAGT against arbitrary-scale SR methods.
\end{itemize}

\noindent\textbf{Practical value.} Beyond its research value, our work offers promising practical applications in copyright protection. In the publishing industry or online image markets, certain images are only available in low resolution for preview. This attack technique ensures that even if someone attempts to recover a high-quality image through arbitrary-scale super-resolution, distortion will still occur, thereby safeguarding the original image’s copyright. Similarly, streaming services (e.g., Netflix) often provide low-resolution previews for non-subscribers. By embedding adversarial perturbations in these low-resolution previews, our approach ensures that any attempts to enhance quality via super-resolution will result in degraded images, thereby preventing unauthorized content reproduction.

\noindent\textbf{Code.} We open-sourcing the code on GitHub at the following URL: \url{https://github.com/Ecnu-luobote/SIAGT}.

\section{Related Work}
\label{sec:relate}
\subsection{Arbitrary-scale Super-resolution}
The single image super-resolution (SISR) \cite{ledig2017photo,lim2017enhanced,zhang2018residual,lai2017deep} on a fixed scale via a deep neural network (DNN) has been well developed. However, most of them limit their upsampling scales to specific integer values and are required to train a distinct model for each upsampling scale. To overcome this limitation, researchers have shifted their focus to arbitrary-scale SR using a unified model. Inspired by implicit neural representation 
\cite{mildenhall2021nerf,sitzmann2019scene,atzmon2020sal,chen2019learning,gropp2020implicit}, LIIF \cite{chen2021learning} first introduced an association of coordinates with continuous images of arbitrary resolutions. Given a query coordinate and deep features of the LR image, the RGB value is predicted via a continuous image function. Subsequently, some LIIF-based arbitrary-scale SR methods have been proposed. For example, UltraSR \cite{xu2021ultrasr} and IPE \cite{liu2021enhancing} utilized positional coding to improve the representation of high-frequency information. LTE \cite{lee2022local} introduced a local texture estimator by Fourier information representation to enhance reconstruction performance. A-LIIF \cite{li2022adaptive} proposed an adaptive local implicit image function for arbitrary-scale super-resolution, enabling continuous representation of high-resolution images by dynamically adjusting to local image characteristics. CiaoSR \cite{cao2023ciaosr} proposed an implicit attention mechanism to learn feature similarities and dynamically generate feature weights. Similarly, CLIT \cite{chen2023cascaded} also utilized the attention mechanism to aggregate the surrounding features. LMF \cite{he2024latent} integrates latent modulated functions with neural field techniques to capture both fine details and global image features. Due to the convenience and flexibility of arbitrary-scale SR in real-world applications, the objective of the adversarial attacks in this study is to unveil vulnerabilities and raise awareness regarding their security.

\subsection{Adversarial Attack on Super-resolution}
SISR is a basic low-level computer vision task that can help generate high-resolution images for HR devices.
Also, the SISR methods have the potential to enhance image quality significantly and are commonly utilized as the key component in various downstream computer vision tasks, including object detection \cite{shermeyer2019effects}, semantic image segmentation \cite{wang2020dual}, and image recognition \cite{fookes2012evaluation}. \textit{Thus ensuring the robustness of image SR is of utmost significance.} 

Adversarial attacks \cite{huang2024TSCUAP,huang2023ALA,jia2020adv,zhang2020interpreting,li2023vrifle,li2023enroll} on classification tasks involve the addition of imperceptible perturbations to input samples, leading to classifiers producing confidently incorrect outputs. Gradient-descent evasion attack \cite{biggio2013evasion,ze2023ultrabd} first studied evasion attacks on machine learning models at test time. Fast Gradient Sign Method (FGSM) then proposed a method to generate adversarial examples based on gradients \cite{goodfellow2014explaining}. However, FGSM did not achieve a high attack success rate due to its one-step attack strategy. To improve the attack performance, Iterative FGSM (I-FGSM) \cite{kurakin2016adversarial} and Projected Gradient Descent (PGD) \cite{madry2017towards} have been proposed to perturb the image with an iterative attack strategy. 

SR is a pixel-level task instead of a feature-level task like classification, and the attack objective is more focused on degrading the image quality rather than compromising the semantic information of SR images. There is still a lack of comprehensive exploration of adversarial attacks against SR. Currently, only Choi et al. \cite{choi2019evaluating} have conducted preliminary explorations of adversarial attack methods for image SR. They evaluated the robustness of fixed-scale SR by optimizing the degradation of the entire SR image output, a factor closely tied to the resolution of the SR image. Consequently, their approach encounters time and memory consumption problems when applied to large-scale SR tasks, rendering it unsuitable for arbitrary-scale SR scenarios.

\subsection{Cross-model Transferability}
In a white-box setting, adversarial attacks require knowing the model architecture and parameters, which are often unknown in practical applications. In the image classification domain, some works \cite{xie2019improving,dong2019evading,lin2019nesterov,gao2020patch,long2022frequency,zhou2018transferable,huang2019enhancing,zhang2022improving,choi2019evaluating} have paid attention to improve the cross-model transferability of the attack. However, to our best knowledge, no paper has explored improving the cross-model transferability of adversarial attacks on SR tasks. Since SR is a pixel-level task that is entirely different from the feature-level task (\eg, classification), improving the cross-model attack transferability on the SR task is an interesting and pioneering problem.

\begin{figure}[tb]
\centering
\includegraphics[width=\linewidth]{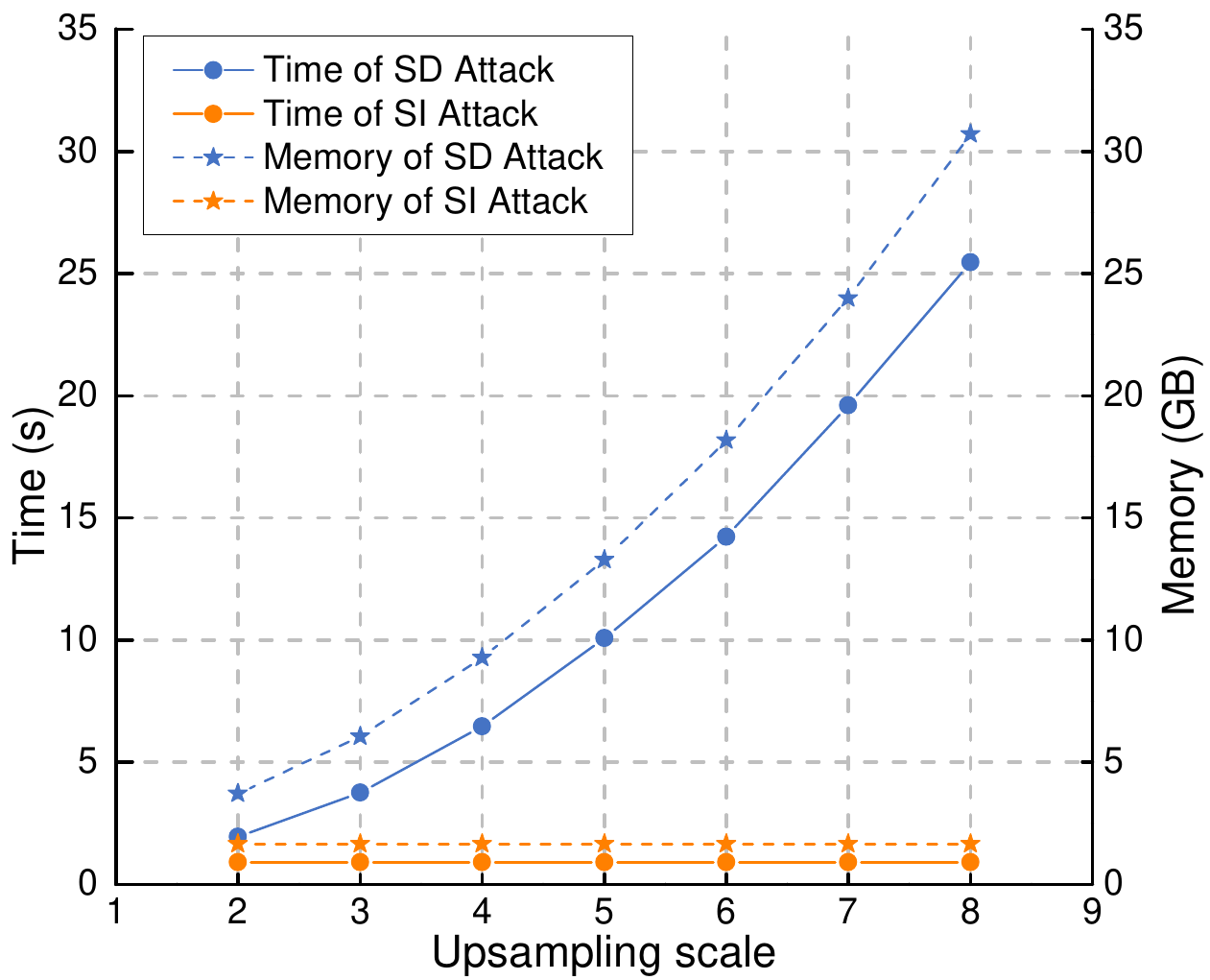}
\caption{Time and memory comparison (on average) between scale-invariant (SI) attack (ours) and scale-dependent (SD) attack \cite{choi2019evaluating} at different scales against LIIF \cite{chen2021learning} on Set5 \cite{bevilacqua2012low}.}
\label{fig:cost_time_memory}
\end{figure}

\section{Attack on Arbitrary-scale Super-Resolution}\label{sec:method}
\subsection{Preliminary}\label{sec:preliminary}
Unlike ``discrete image representation'' used in fixed-scale SR methods which represents images discretely with 2D arrays of pixels, inspired by the continuous nature of the visual world, ``continuous image representation'' (in this paper the term ``continuous representation'' is the same as ``continuous image representation'' since the task is single image SR) is a popular new concept \cite{chen2021learning} used in arbitrary-scale SR methods that proposed to \textbf{denote images as a continuous function} where pixel values are calculated by using the corresponding coordinate points as inputs (coordinate points are continuous). 
The image expressed by a continuous representation is referred to as a continuous image.

To be specific, given an LR image $I^{LR} \in \mathbb{R}^{H \times W \times 3}$ and an arbitrary upsampling scale $\mathbf{r}= \{r_h, r_w\}$, where $r_h$ and $r_w$ denote any proper real-number scale factors, arbitrary-scale SR method aims to achieve an SR image $I^{SR} \in \mathbb{R}^{r_{h}H \times r_{w}W \times 3}$. 
Currently, LIIF-based SR methods \cite{chen2021learning,xu2021ultrasr,liu2021enhancing,lee2022local,cao2023ciaosr} are the most popular arbitrary-scale SR methods and show SOTA SR performance benefited from their continuous representation design. The core of LIIF-based SR methods is two functions: a DNN encoder $E_{\varphi}(\cdot)$ and a continuous image function $f_{\theta}(\cdot)$ (usually parameterized as a multi-layer perceptron (MLP)). The DNN encoder $E_{\varphi}(\cdot)$ extracts latent code (\ie, feature) $\mathcal{Z} = E_{\varphi}(I^{LR})$ from $I^{LR}$, and the continuous image function $f_{\theta}$ takes both the latent code $\mathcal{Z}$ and coordinates as inputs and generate RGB values. Formally, the continuous image function $f_{\theta}(\mathcal{Z},x): \mapsto \mathcal{S}$ can be defined as follows,
\begin{equation}
  s = f_{\theta} (\mathcal{Z},x),
  \label{eq:1}
\end{equation}
where $x \in \mathcal{X}$ is a 2D coordinate in the continuous image domain, and $s \in \mathcal{S}$ is the predicted RGB value of $I^{SR}$. We can find that for the LIIF-based methods, calculating each RGB value of the pixel in $I^{SR}$ needs one query (each query means performing Eq. \eqref{eq:1} once with a specific coordinate point as input) and totally needs ($r_{h}H \times r_{w}W$) queries for the generation of $I^{SR}$, which is significantly time-consuming. Please note that continuous image $I$ is unique, $I^{LR}$ and $I^{SR}$ can be seen as discrete representations of the continuous image $I$ under different resolutions. The latent code $\mathcal{Z}$ is \textit{not} continuously represented while $s$ is continuously represented since $s$ is calculated with coordinate $x$ (coordinate points are continuous).

\subsection{Problem Formulation and Motivation}\label{sec:problem_formulation_motivation}
\noindent\textbf{Attack objective.} Adversarial attacks against the SISR method refer to artificially injecting small human-imperceptible perturbations into the $I^{LR}$ image, with the purpose of generating a significantly deteriorated $I^{SR}$ image. To be specific, given the LR image $I^{LR}$, we can add an adversarial perturbation $\delta$ on it to obtain the adversarial LR image $I^{adv} = I^{LR}+\delta$. The optimization objective is
\begin{align}\label{eq:pgd}
\delta = \argmax_{\delta^{*}}{\mathcal{L}(I^{adv}, I^{LR}),~\text{subject to}~\|\delta^{*}\|_p\leq\epsilon},
\end{align}
where $\mathcal{L}(\cdot)$ denotes the loss function and $\epsilon$ is the attack strength which restricts the adversarial perturbation $\delta$ to be small. 

\noindent\textbf{Problem of scale-dependent attack.} The previous SR adversarial attack methods are scale-dependent (SD) attack methods. They attempt to maximize the amount of deterioration on the discrete representation SR image generated from $I^{LR}$ by a fixed upsampling scale $\mathbf{r}$. Specifically, let $\Phi(\cdot,\cdot)$ denotes the fixed-scale SR model, the previous SR adversarial attack method \cite{choi2019evaluating} uses the $L_2$ distance between the SR output $\Phi(I^{LR},\mathbf{r})$ of the original clean LR image and the SR output $\Phi(I^{adv},\mathbf{r})$ of the adversarial LR image as the loss function
\begin{equation}
\mathcal{L_{SD}} = \|\Phi(I^{adv},\mathbf{r}) -\Phi(I^{LR},\mathbf{r})\|_{2}.
\label{eq:scale-dependent_loss}
\end{equation}
This method shows outstanding attack performance against fixed-scale SR methods.

With the flourishing of LIIF-based SR methods due to their convenient and efficient practical benefits, adversarial attacks for evaluating the vulnerability of arbitrary-scale SR methods are necessary. However, scale-dependent attack methods may not be feasible with arbitrary-scale SR and show terrible behavior in practical applications. Obviously, since each RGB value of the pixel in $I^{SR}$ needs one query and totally needs ($r_{h}H \times r_{w}W$) queries for the generation of whole $I^{SR}$, as well as the scale-dependent attack needs the discrete representation SR image $I^{SR}$ to construct the loss, thus with the increase of upsampling scale \textbf{r}, the time and resource consumption of scale-dependent attack will gradually become huge and unacceptable. As shown in Figure~\ref{fig:cost_time_memory}, we employ the scale-dependent attack \cite{choi2019evaluating} on LIIF under multiple upsampling scales on super-resolution dataset Set5 \cite{bevilacqua2012low} with an Nvidia A6000 GPU of 48GB RAM to show the time and memory consumption. The consumption of the scale-dependent attack is in blue. We can find that, with the increase in scale, the time and memory cost of the attack show \textit{exponential rise}, which is a catastrophic behavior. This raises the question of designing a scale-invariant attack for the arbitrary-scale SR. 
Obviously, the attack \textit{should be applied on the continuous image rather than on the discrete SR image} since this can ensure the deterioration of the SR image at any upsampling scale due to the fact that the generation of SR image at any scale needs the continuous image as input.
However, facing an image representation that is continuous (\ie, can be infinitely subdivided and has infinite coordinates in the continuous representation domain), it is hard to construct loss like Eq.~\eqref{eq:scale-dependent_loss} since it is impossible to traverse all coordinate points. Furthermore, we also verify that discarding coordinates and only leveraging the latent code $\mathcal{Z}$ to construct loss will lead to obviously lower attack performance than the scale-dependent SR attack method (see Section~\ref{sec:discussion} and Table~\ref{tab:baseline_comparision}). 
Therefore, how to design an efficient and effective scale-invariant adversarial attack for arbitrary-scale SR is an important and thorny challenge. 

\noindent\textbf{High-level idea.}
A simple and elegant idea is to \textit{alter the continuous image by adjusting the values of finite discrete coordinate points in it}. This is inspired by an intuitive thinking that, changing the value of a single coordinate point in the continuous image will cause the value of many coordinate points to change.   
For instance, consider a continuous image that has point A at coordinate (0, 0) with a value of 0, surrounded by adjacent points also valued at 0. If we change A's value to 1, the values of the adjacent points will also change to become non-zero values to satisfy the continuity in the representation.
On the other hand, we also find that the influence range (\ie, number of the influenced coordinate points) of changing A's value is large.
Therefore, by modifying a finite number of discrete (\ie, non-adjacent) points of the continuous image, we can alter the value of a large number of coordinate points in the continuous image, thus efficiently constructing attacks to achieve the deteriorated continuous image.   

\begin{figure}[tb]
\centering
\includegraphics[width=\linewidth]{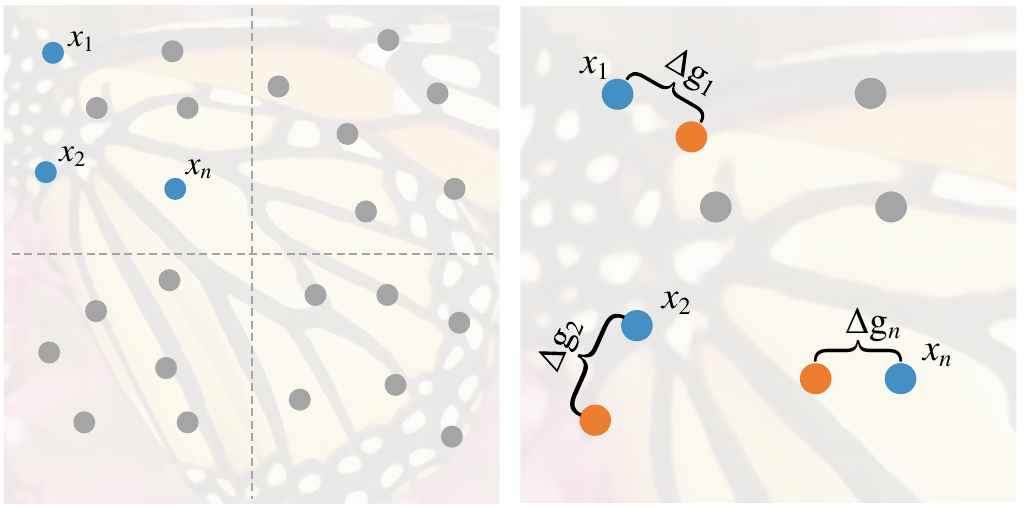}
\caption{Diagram of SIAGT on continuous image $I$. (L) Divide $I$ into 4 blocks and query $n$ coordinates (blue point) per block. (R) To improve the attack transferability on arbitrary-scale SR task, we propose to expand the value gap between the queried coordinate ($x_i$) (blue point) with its neighboring coordinate ($x_i + \Delta g_{i}$) (orange point) to construct $L_{TR}$ loss. Here we use the top-left portion of the (L) figure to illustrate the queried coordinates and their neighboring coordinates.}
\label{fig:method}
\end{figure}

Based on this high-level idea, our proposed scale-invariant attack method (in orange of Figure~\ref{fig:cost_time_memory}) shows excellent time-saving and memory-saving, basically invariant with scale. The detailed design of our method is presented in Section~\ref{sec:attack}.

\subsection{Scale-invariant Adversarial Attack}\label{sec:attack}
Following the high-level idea, we suggest constructing the \textbf{scale-invariant (SI) attack} on the continuous image instead of discrete SR output. The attack on the continuous image equals simultaneously attacking to SR images at arbitrary scales. To be specific, the proposed method has two steps. First, we divide the continuous image $I$ into finite blocks. Second, we construct the loss with RGB values calculated by randomly selected coordinates in each block. In our empirical study, we find that evenly splitting the continuous image $I$ into $H \times W$ blocks can achieve excellent attack performance (see Table~\ref{tab:attack_result}). Obviously, the complexity of the scale-invariant and scale-dependent attack method is $O(H \times W)$ and $O(r_{h}H \times r_{w} W)$. This means the time and memory consumption of the scale-invariant attack is \textit{fixed} while the consumption of the scale-dependent attack \textit{increases exponentially} with the upsampling scale. Thus, if the upsampling scale $\mathbf{r} = \{30,30\}$ (a common large scale in arbitrary-scale SR tasks), our method is hundreds of times faster than the scale-dependent SR attack \cite{choi2019evaluating}. 

To describe the method in detail, as an example, in Figure \ref{fig:method} (L), we divide the continuous image $I$ into $2 \times 2$ ($H=2$, $W=2$) blocks. The gray dots are coordinates under continuous representation in the block. In our attack method, we randomly query $n$ coordinates (\ie, $x_{1},x_{2},\cdots,x_{n}$) per block and the loss $\mathcal{L}_{SI}$ is defined as:
\begin{equation}
\mathcal{L}_{SI} = \|\Psi( I^{adv},\Lambda)-\Psi( I^{LR},\Lambda)\|_{2}, \\
\label{eq:loss_SI}
\end{equation}
where $\Lambda$ denotes the set of query coordinates of all blocks, $\Psi(\cdot,\cdot)$ denotes the arbitrary-scale SR model. Here we use  $\Psi(\cdot,\cdot)$ to represent arbitrary-scale SR model for distinguishing from fixed-scale SR model $\Phi(\cdot,\cdot)$. Please note that the $\mathcal{L}_{SI}$ loss can not be directly applied to the selected points in latent code $\mathcal{Z}$ generated by the encoder $E_{\varphi}(\cdot)$ since the latent code $\mathcal{Z}$ is not under continuous representation and only the coordinate $x \in \mathcal{X}$ is in the continuous image domain. Thus here we use the finite RGB values in SR results of $I^{LR}$ and $I^{adv}$ to reflect the information of \textit{finite discrete points in continuous image} since they are generated with the given latent code $\mathcal{Z}$ and coordinates.

\subsection{Improving Attack Transferability}\label{sec:transfer_attack}
Cross-model transferability is important for white-box attacks since it can help researchers discover similar weaknesses or vulnerabilities in models and save computing resources when attacking multiple models, here we further study the cross-model transferability of attack on arbitrary-scale SR. Existing methods focus on the universality and invariance of the feature to improve the cross-model transferability, which fits the classification tasks that are strongly feature-dependent but incompatible with the arbitrary-scale SR task that is coordinate-dependent. To solve the \textit{new research problem of improving the attack transferability on SR task based on continuous representation}, we propose to design the method by focusing on coordinates. 

\begin{figure}[tb]
\centering
\includegraphics[width=\linewidth]{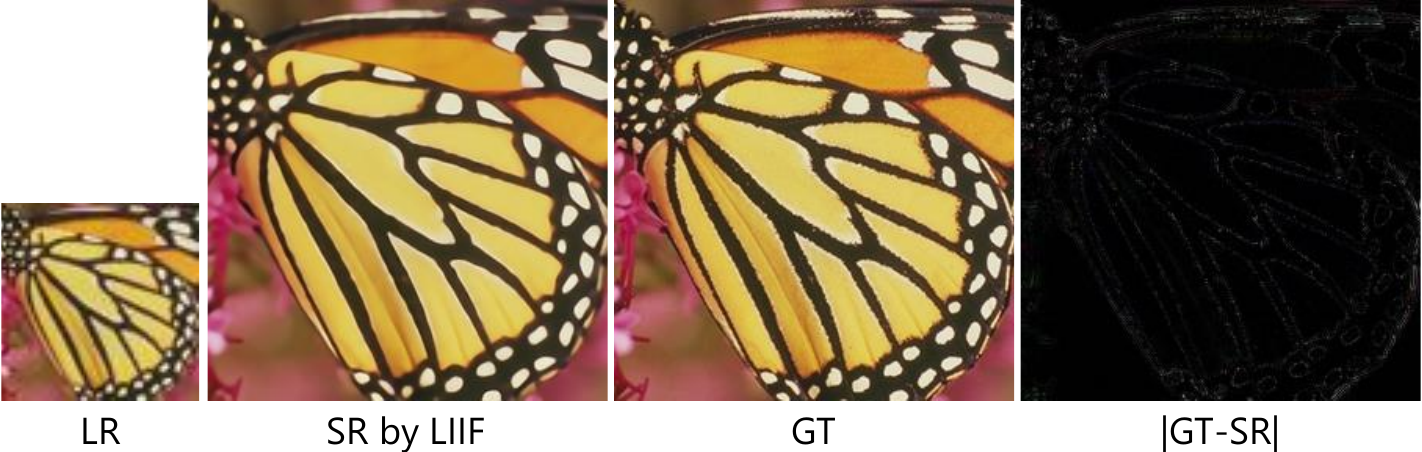}
\caption{Performance of LIIF on handling low-frequency and high-frequency information in the image.}
\label{fig:difference_between_GT_and_SR_high_frequency}
\end{figure}

Our opinion is that \textbf{increasing high-frequency information in adversarial low-resolution images degrades the performance of arbitrary super-resolution (ASSR) models by making restoration more difficult.} The high-level idea is based on the observation that super-resolution models can handle low-frequency information (content includes smooth, gradually varying areas of the image, typically seen in large regions with minimal color or brightness variation) in images relatively easily, while high-frequency information (content includes detailed, rapidly changing areas in the image, such as edges, textures, and noise) remains challenging to reconstruct effectively. This issue is widely acknowledged in SR literature (e.g., \cite{liang2022details,xie2021learning}). We also evaluate the performance of an arbitrary-scale method (i.e., LIIF) on low-resolution images to assess the effect on both low- and high-frequency components. In Figure~\ref{fig:difference_between_GT_and_SR_high_frequency} list the low-resolution input, the LIIF super-resolved output (SR), its high-resolution ground truth (GT), and the difference between GT and SR. Discrepancies predominantly occur in high-frequency regions (e.g., edges, textures), underscoring the inherent difficulty in reconstructing high-frequency information. This inherent limitation of SR models is well-documented in SR-related literature (e.g., \cite{liang2022details, xie2021learning}) and also confirmed by our empirical experiments (Figure~\ref{fig:difference_between_GT_and_SR_high_frequency}), where discrepancies between ground truth and super-resolved images are pronounced in high-frequency regions. By exploiting this weakness of SR models, our method deliberately increases high-frequency information in adversarial low-resolution images, making it substantially harder for SR models to generate high-quality outputs. Since this limitation is shared across different SR models, increasing high-frequency information improves the cross-model attack transferability of our method. This exploitation of SR models’ difficulty with high-frequency reconstruction forms the foundation of our approach.

Following the above idea, we propose to increase the high-frequency information in the adversarial image. Specifically, we aim to expand the gap of RGB values between the query coordinate and its neighboring coordinates (see Figure~\ref{fig:method} (R)). Formally, the loss $\mathcal{L}_{TR}$ expressed as
\begin{equation}
\mathcal{L}_{TR} = \|\Psi(I^{adv},\Lambda \oplus \Delta G)-\Psi(I^{adv},\Lambda)\|_{2},
\label{eq:loss_TR}
\end{equation}
where $\Delta G$ is a set of randomly generated coordinate offsets. $\oplus$ means selecting the corresponding coordinates in two sets ($\Lambda$ and $\Delta G$) with the same index and plus them together. Note that although our approach aims to maximize value differences between adjacent coordinates in the continuous image with Eq.~\eqref{eq:loss_TR}, it still adheres to the perturbation constraint Eq.~\eqref{eq:pgd}, resulting in a nearly imperceptible attack. The operation of maximizing value differences between adjacent coordinates does not target specific RGB values or colors, such as contrasting colors, for neighboring coordinates. With the straightforward optimization objective (reducing the quality of the super-resolved image and increasing RGB differences between neighboring coordinates) defined by Eq.~\eqref{eq:loss_SI} and Eq.~\eqref{eq:loss_TR}, the perturbations can be generated automatically through gradient-based adjustments derived from the loss function.
Intuitively, the larger value difference between adjacent coordinates in the continuous image will lead to larger pixel differences in the $I^{adv}$ image, thus generating more high-frequency information in the $I^{adv}$ image. For any arbitrary-scale SR model, super-resolution of high-frequency information is difficult, so including $\mathcal{L}_{TR}$ loss can improve the cross-model transferability of the attack. 
The total loss function of our SIAGT method is:
\begin{equation}
\mathcal{L}_{final} = \mathcal{L}_{SI} + \beta \cdot \mathcal{L}_{TR},
\label{eq:total_loss}
\end{equation}
where $\beta$ is the weight value for balancing the ${L}_{SI}$ and ${L}_{TR}$.

\begin{table*}[tb]
\centering
\caption{The attack performance (PSNR (dB)) of SIAGT on arbitrary-scale SR models at different upsampling scales, where the `LR' column represents the PSNR value between the original input image $I^{LR}$ and the adversarial image $I^{adv}$.}
\begin{tabular}{l|l|c|ccccccccc}
\hline 
Dataset & Model & LR & $\times 2$  & $\times 3$ & $\times 4$  & $\times 6$  & $\times 8$  & $\times 12$  & $\times 18$  & $\times 24$ & $\times 30$ \tabularnewline
\hline 
\multirow{3}{*}{Set5 \cite{bevilacqua2012low}}
& LIIF \cite{chen2021learning} & 37.87 & 20.34 & 19.07 & 19.05 & 19.24 & 19.41 & 19.63 & 19.80 & 19.89 & 19.95\tabularnewline
& LTE \cite{lee2022local}     & 37.87 & 20.98 & 19.84 & 19.63 & 19.97 & 20.52 & 21.33 & 21.94 & 22.25 & 22.44\tabularnewline
& CiaoSR \cite{cao2023ciaosr} & 32.44 & 14.76&	14.67 &13.98	&13.96	&14.15&	14.09&14.13	&14.17	&14.23
\tabularnewline
\hline 
\multirow{3}{*}{Set14 \cite{zeyde2012single}}
& LIIF & 38.22 & 18.70 & 17.44 & 17.50 & 17.74 & 17.98 & 18.31 & 18.53 & 18.64 & 18.71\tabularnewline
& LTE & 38.27 & 19.50 & 18.33 & 18.32 & 18.88 & 19.53 & 20.41 & 21.31 & 21.66 & 21.87\tabularnewline
& CiaoSR & 32.68 & 13.42&13.60&12.80&12.82&12.93&13.04&13.18&13.19 &13.25
\tabularnewline
\hline 
\multirow{3}{*}{B100 \cite{martin2001database}} 
& LIIF & 38.36 & 17.85 & 16.56 & 16.62 & 16.85 & 17.07 & 17.37 & 17.56 & 17.69 & 17.75\tabularnewline
 & LTE& 38.49 & 18.25 & 17.12 & 17.14 & 17.73 & 18.37 & 19.26 & 19.90 & 20.23 & 20.42\tabularnewline
 & CiaoSR& 32.80 & 12.97 &13.11&12.35& 12.56&12.58&12.73&12.78&12.84&12.89
 \tabularnewline
\hline
\multirow{3}{*}{Urban100 \cite{huang2015single}} 
& LIIF
& 38.11 & 17.41 & 16.10 & 16.15 & 16.42 & 16.66 & 16.96 & 17.12 & 17.29 & 17.36\tabularnewline
& LTE
& 38.26 & 18.01 & 16.78 & 16.83 & 17.61 & 18.42 & 19.49 & 20.07 & 20.67 & 20.91\tabularnewline
& CiaoSR
& 32.78 & 12.90 & 12.81 &12.22 & 12.30&12.40 &12.53&12.61&12.66 &12.69\tabularnewline
\hline
\end{tabular}
\label{tab:attack_result}
\end{table*}
\begin{figure*}[tb]
\centering
\includegraphics[width=\linewidth]{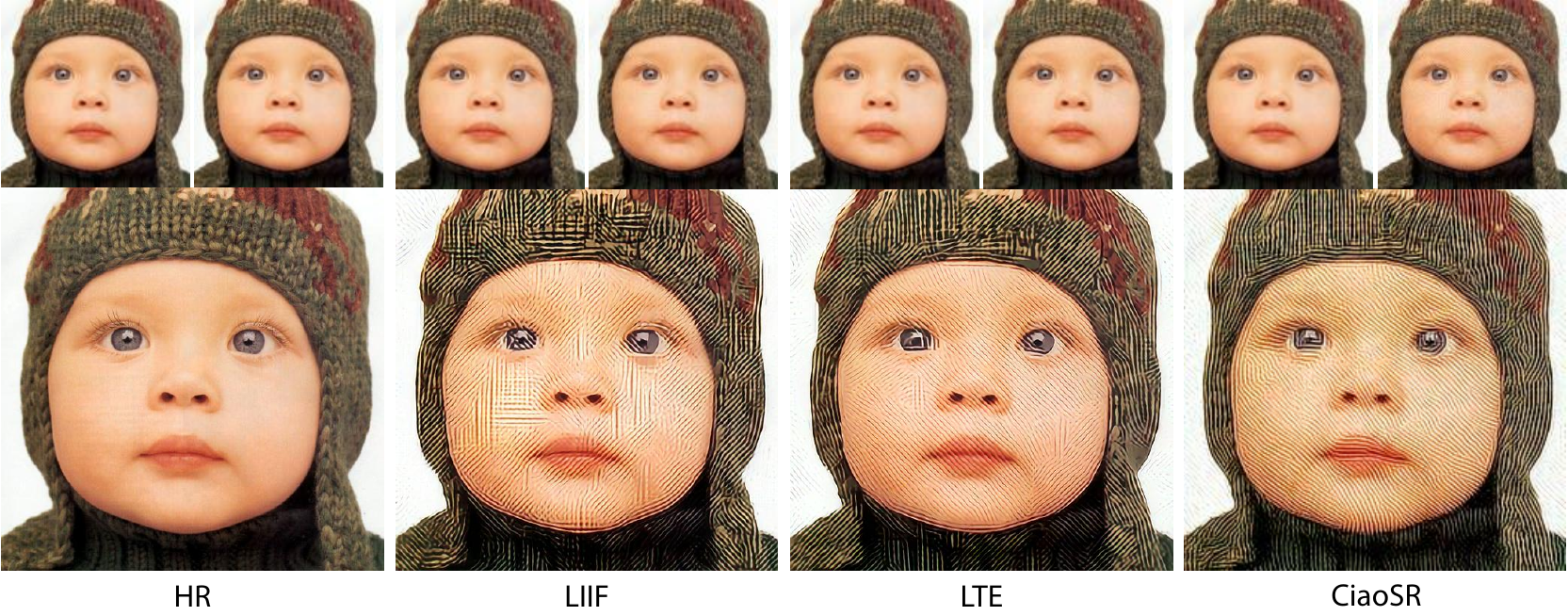}
\caption{Visualization of the attack results by different SR models. (top-left) is the original input clean image, (top-right) is the adversarial image, and (bottom) is the SR output ($\times 4$) obtained from the adversarial images.}
\label{fig:robustness}
\end{figure*}
\begin{figure}[tb]
\centering
\includegraphics[width=0.8\linewidth]{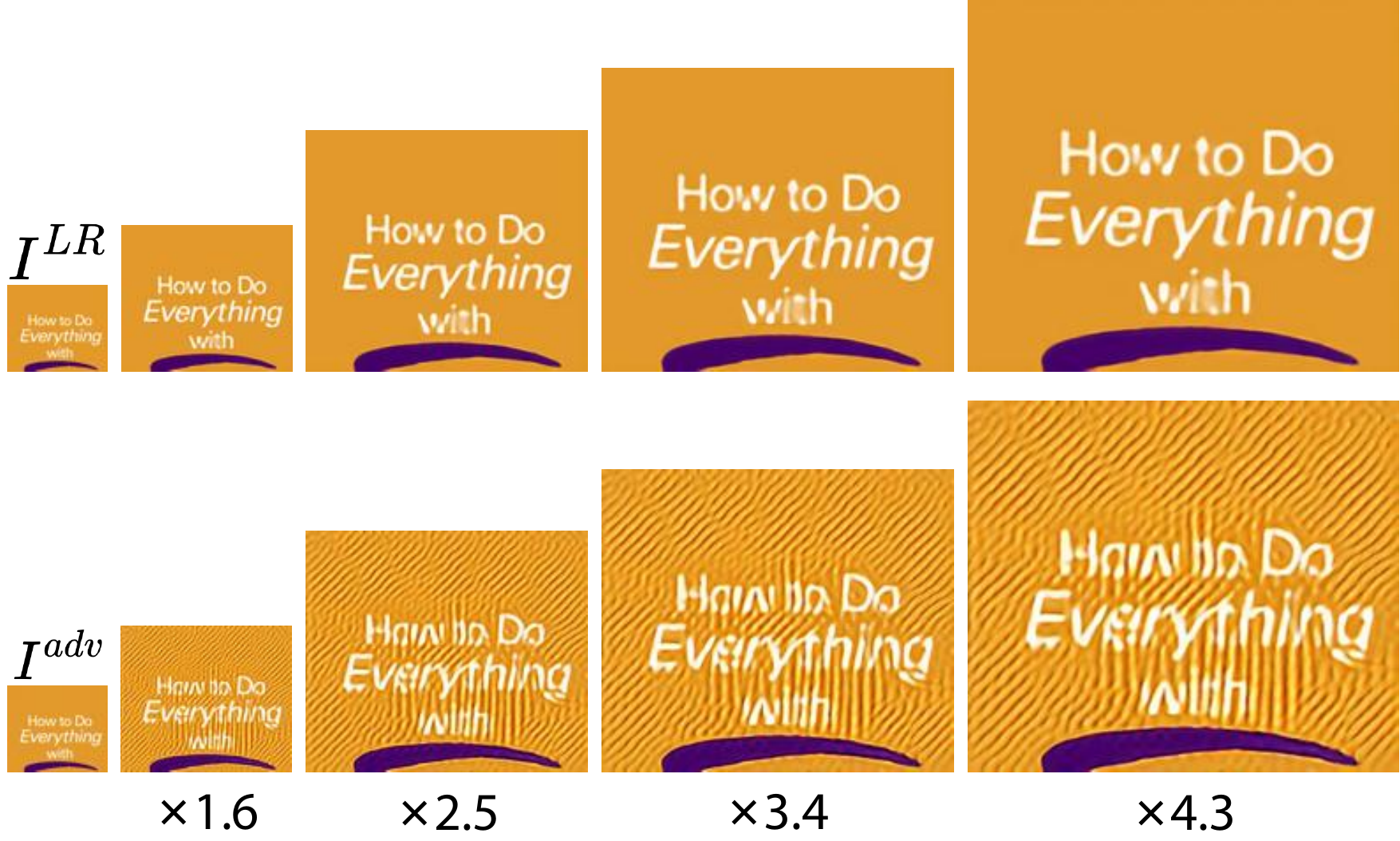}
\caption{Visualization of the attack results with non-integer scale factors generated from $I^{LR}$ (top) and $I^{adv}$ (bottom).}
\label{fig:multi_scale}
\end{figure}

\subsection{Attack Iteration Strategy}
For the iteration strategy, we refer to commonly used PGD-like $l_\infty$-norm attacks except for modification on perturbation magnitude in each iteration. To be specific, the strategy for updating $I^{adv}$ (in the implementation of LIIF, the value of $I^{adv}$ is in the range [-1,1]) with traditional PGD is:
\begin{equation}
    \Tilde{I}_{t+1}^{adv}= \mathrm{clip}_{-1,1}\Big(I_{t}^{adv} + \frac{\epsilon}{T} ~\mathbf{sign}~\big(\nabla \mathcal{L}_{final}\big)\Big),   \label{eq:attack_strategy_clip_image} 
\end{equation}
\begin{equation}
    I_{t+1}^{adv} = \mathrm{clip}_{-\epsilon,\epsilon}(\Tilde{I}_{t+1}^{adv} -I^{LR}) + I^{LR},
\label{eq:attack_strategy_clip_eps}
\end{equation}
where $\epsilon$ is the attack strength, $t$ is the index of the current iteration in the range [1, $T$], $T$ is the total number of iterations, $\mathbf{sign}~\big(\nabla \mathcal{L}_{final}\big)$ is the sign of the gradient, and 
\begin{gather}
    \mathrm{clip}_{a,b}(I) = \min(\max(I,a),b).
\label{eq:clip_function}
\end{gather}
However, we empirically find that using fixed-strength $\epsilon/T$ per iteration does not achieve a good attack performance (see Table~\ref{tab:iteration_ablation}). Thus we propose replacing Eq.~\eqref{eq:attack_strategy_clip_image} with
\begin{equation}
    \Tilde{I}_{t+1}^{adv}= \mathrm{clip}_{-1,1}\Big(I_{t}^{adv} + \frac{\epsilon}{t} ~\mathbf{sign}~\big(\nabla \mathcal{L}_{final}\big)\Big).
    \label{eq:attack_strategy_clip_image_new}
\end{equation}
This operation aims to start with a large perturbation amplitude per iteration and make it smaller as the number of iterations increases. The strategy helps approximate adversarial examples in a faster manner than Eq.~\eqref{eq:attack_strategy_clip_image}.

\section{Experiment}\label{sec:experiment}
\noindent\textbf{Datasets.} We use four widely used classical benchmark datasets for evaluation, including Set5 \cite{bevilacqua2012low}, Set14 \cite{zeyde2012single}, B100 \cite{martin2001database}, and Urban100 \cite{huang2015single}. These datasets consist of 5, 14, 100, and 100 images, respectively. We employ the bicubic downsample on images for generating the input LR images.

\noindent\textbf{Target model.} We choose the three arbitrary-scale SR models: the foundational one (LIIF \cite{chen2021learning}) and two classical models (LTE \cite{lee2022local}, CiaoSR \cite{cao2023ciaosr}), as the target SR model and use pre-trained weight provided by official sources.

\begin{figure*}[tb]
\centering
\includegraphics[width=\linewidth]{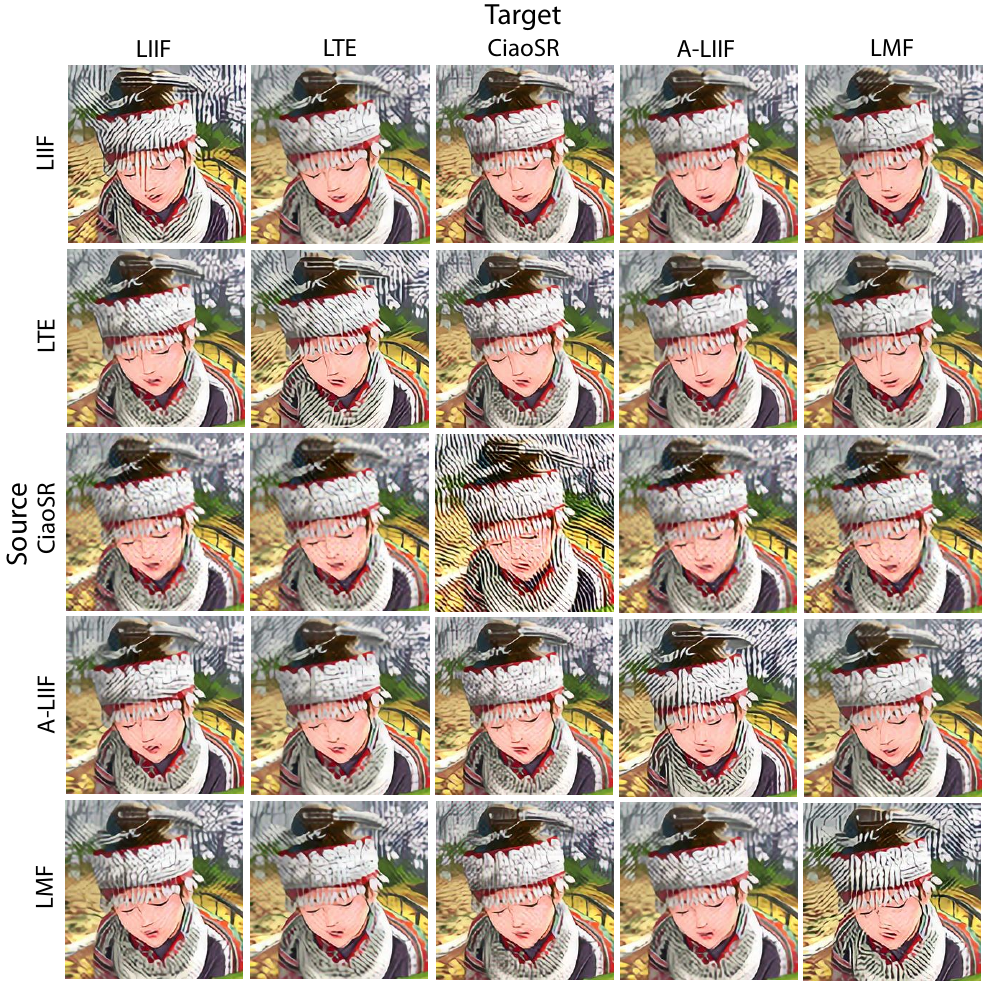}
\caption{The visual representation of cross-model attack transferability with $\mathcal{L}_{final}$ loss.}
\label{fig:cross-model}
\end{figure*}
\begin{table*}[tb]
\centering
\setlength{\tabcolsep}{3pt}
\caption{Evaluating cross-model attack transferability by using $\mathcal{L}_{final}$ and $\mathcal{L}_{SI}$ on attack across five different SR models.}
\resizebox{\linewidth}{!}{
\begin{tabular}{l|c|c|ccccc|ccccc|ccccc|ccccc|ccccc}
\hline
\multicolumn{1}{c|}{\multirow{3}{*}{}} & \multicolumn{1}{c}{\multirow{3}{*}{}} & \multicolumn{1}{c|}{\multirow{4}{*}{}} & \multicolumn{25}{c}{Target}\tabularnewline
\cline{4-28}
 Method & \multicolumn{2}{c|}{SR Model}& \multicolumn{5}{c|}{LIIF} & \multicolumn{5}{c|}{LTE} & \multicolumn{5}{c|}{CiaoSR}& \multicolumn{5}{c|}{A-LIIF}& \multicolumn{5}{c}{LMF}\tabularnewline
 &  \multicolumn{1}{c}{} &  & $\times 2$ & $\times 4$ & $\times 8$ & $\times 18$ & $\times 30$ & $\times 2$ & $\times 4$ & $\times 8$ & $\times 18$ & $\times 30$ & $\times 2$ & $\times 4$ & $\times 8$ & $\times 18$ & $\times 30$ & $\times 2$ & $\times 4$ & $\times 8$ & $\times 18$ & $\times 30$& $\times 2$ & $\times 4$ & $\times 8$ & $\times 18$ & $\times 30$\tabularnewline
\hline 

$\mathcal{L}_{SI}$ & \multirow{10}{*}{Source} & \multirow{2}{*}{LIIF} & 17.91 & 16.70 & 17.08 & 17.47 & 17.63 & 26.44 & 25.42 & 25.90 & 26.73 & 27.05 & 25.18 & 23.73 & 23.62 & 23.66 & 23.68 &28.19&27.13&26.82&26.71&26.88&25.80&24.35&24.26&24.41&24.49\tabularnewline
$\mathcal{L}_{final}$ &  &  & 17.85 & 16.62 & 17.07 & 17.56 & 17.75 & \textbf{26.08} & \textbf{24.88} & \textbf{25.43} & \textbf{26.35} & \textbf{26.69} & \textbf{24.74} & \textbf{23.20} & \textbf{23.09} & \textbf{23.14} & \textbf{23.15}& \textbf{27.90}& \textbf{26.69} & \textbf{26.39} & \textbf{26.30}& \textbf{26.28}&\textbf{25.21}& \textbf{23.96} & \textbf{23.93} &\textbf{24.13}& \textbf{24.23} \tabularnewline
\cline{1-1} \cline{3-28} 
$\mathcal{L}_{SI}$ &  & \multirow{2}{*}{LTE} & 26.88 & 25.36 & 25.18 & 25.29 & 25.35 & 18.34 & 17.29 & 18.37 & 19.76 & 20.25 & 24.20 & 22.62 & 22.46 & 22.49 & 22.53&28.36&27.27&27.00&26.90&26.88&25.49&24.05&24.04&24.25&24.34 \tabularnewline
$\mathcal{L}_{final}$ &  &  & \textbf{26.71} & \textbf{25.17} & \textbf{25.03} & \textbf{25.19} & \textbf{25.29} & 18.25 & 17.14 & 18.37 & 19.90 & 20.42 & \textbf{23.93} & \textbf{22.25} & \textbf{22.09} & \textbf{22.13} & \textbf{22.15} & \textbf{28.36} & \textbf{27.17} & \textbf{26.89} & \textbf{26.81}& \textbf{26.80} & \textbf{25.43} & \textbf{23.84} & \textbf{23.90}& \textbf{24.16}& \textbf{24.27}\tabularnewline
\cline{1-1} \cline{3-28} 
$\mathcal{L}_{SI}$ &  & \multirow{2}{*}{CiaoSR} & 28.43 & 27.57 & 27.37 & 27.33 & 27.33 & 28.39 & 27.64 & 27.89 & 28.39 & 28.58 & 12.91 & 12.26 & 12.36 & 12.51 & 12.60&28.69& 27.89& 27.61& 27.48& 27.44& 27.86&26.68&26.42&26.44&26.47\tabularnewline
$\mathcal{L}_{final}$ &  &  & \textbf{28.05} & \textbf{27.01} & \textbf{26.82} & \textbf{26.81} & \textbf{26.83} & \textbf{28.11} & \textbf{27.21} & \textbf{27.43} & \textbf{27.97} & \textbf{28.17} & 12.97 & 12.35 & 12.58 & 12.78 & 12.87&  \textbf{28.42} &  \textbf{27.49} &\textbf{27.18} 
 & \textbf{27.05} & \textbf{27.02} &\textbf{27.42} & \textbf{26.08} &\textbf{25.85} & \textbf{25.91} & \textbf{25.96} \tabularnewline
\cline{1-1} \cline{3-28} 
$\mathcal{L}_{SI}$ &  & \multirow{2}{*}{A-LIIF} & 26.46 & 25.32 &  25.12 & 25.15 &25.18 & 27.17 & 26.21 & 26.66 & 27.41 & 27.69 & 24.40 & 23.15  & 22.92 & 22.82 & 22.79 &18.91& 17.70 & 17.52& 17.49& 17.50 &26.09&24.73&24.63&24.75 &24.82\tabularnewline
$\mathcal{L}_{final}$ &   &  &\textbf{26.14}  & \textbf{24.83} &\textbf{24.69}  & \textbf{24.78} & \textbf{24.83} &\textbf{26.99}&\textbf{25.87} & \textbf{26.34} & \textbf{27.15} & \textbf{27.45} & \textbf{24.33} & \textbf{22.88} & \textbf{22.63} & \textbf{22.53} & \textbf{22.49} 
& 18.74& 17.40 & 17.24 & 17.22& 17.23
& \textbf{25.87}
& \textbf{24.37}
& \textbf{24.31}
& \textbf{24.48}
& \textbf{24.57}\tabularnewline
\cline{1-1} \cline{3-28} 
$\mathcal{L}_{SI}$ &  & \multirow{2}{*}{LMF} & 26.38 &25.20  &25.03  &25.08  & 25.12 & 25.87 & 24.80 &25.27 &26.08 &26.40 &24.01  & 22.82 &22.69  &22.62  &22.59& 28.48&27.47&27.15&27.01&26.98&17.53&16.24&16.63&17.10&17.28\tabularnewline
$\mathcal{L}_{final}$ & & & \textbf{26.34} & \textbf{25.16} & \textbf{24.99} & \textbf{25.05} & \textbf{25.08} & \textbf{25.82} & \textbf{24.68} & \textbf{25.17} & \textbf{26.02} & \textbf{26.34} & \textbf{23.90} & \textbf{22.63} & \textbf{22.46} & \textbf{22.38} & \textbf{22.35} & \textbf{28.46} & \textbf{27.44}&\textbf{27.11} &\textbf{26.98} & \textbf{26.95} &17.49&16.13&16.55&17.05&17.24\tabularnewline
\hline 
\end{tabular}
}
\label{tab:cross_model_result}
\end{table*}

\noindent\textbf{Evaluation metrics.}
PSNR (dB) \cite{hore2010image} is the main evaluation metric of arbitrary-scale SR task, and the previous SR attack method \cite{choi2019evaluating} also only uses the PSNR to evaluate the attack performance, thus we follow their setting. A higher PSNR means a better image quality. An ideal SR attack should keep high PSNR between the LR clean ($I^{LR}$) and adversarial ($I^{adv}$) images while leading to low PSNR between their corresponding super-resolved ones. 

\noindent\textbf{Implementation details.}
We evaluate the robustness of arbitrary-scale SR against the attack strength of $\epsilon \leq 8/255$ (a common selection for $l_{\infty}$-norm attacks).  Following the setting in arbitrary-scale SR \cite{chen2021learning,lee2022local,cao2023ciaosr}, the upsampling scale $s \in \{2, 3, 4, 6, 8, 12, 18, 24, 30\}$, which can illustrate as fully as possible the effect of the attack method at different scales. We set the range of $\Delta G$ as $[-0.5 \times 10^{-2},0.5 \times 10^{-2}]$ and $\beta =2$. Our experiments are all performed on an Ubuntu system with Nvidia A6000 GPUs.

\subsection{Attack Performance}
Table~\ref{tab:attack_result} shows the attack performance of SIAGT on LIIF, LTE, and CiaoSR across four datasets. In the LR column, it shows the PSNR between $I^{LR}$ and $I^{adv}$, while other columns (upsampling scale) show the PSNR between their corresponding SR images. We can find that $I^{LR}$ and $I^{adv}$ are very similar (PSNR $>$ 30), which can be regarded as visually identical images \cite{huang2011robust}. For SR images of different scales, most PSNR values are close to or below 20, which demonstrates the scale-invariance and effectiveness of the SIAGT method. 
Figure~\ref{fig:robustness} visualizes SR images ($\times 4$) of $I^{LR}$ and $I^{adv}$. It can be observed that there is no obvious visual difference between $I^{LR}$ and $I^{adv}$, while the SR images of $I^{adv}$ are deteriorated. In addition, we conduct experiments at non-integer scale factors with $I^{LR}$ and $I^{adv}$. In Figure \ref{fig:multi_scale},
the character in the SR images of $I^{adv}$ exhibit significant deformation and distortion, exposing the potential vulnerability of employing arbitrary-scale SR as a pre-processing technique. Furthermore, we explore the influence of super-resolved adversarial low-resolution images on downstream tasks such as text recognition, face recognition and number recognition in Discussion Section~\ref{sec:discussion}.

\begin{table}[tb]
\centering
\caption{Time consumption across different arbitrary-scale approaches and datasets.}
\begin{tabular}{l|ccc}
\hline 
\multicolumn{1}{l|}{Time (s)} & LIIF & LTE & CiaoSR\tabularnewline
\hline 
Set5 & 3.64 & 4.50 & 12.65\tabularnewline
Set14 & 7.46 & 9.36 & 27.56\tabularnewline
B100 & 4.57 & 5.77 & 16.99\tabularnewline
Urban100 & 5.62 & 7.06 & 20.80\tabularnewline
\hline 
\end{tabular}
\label{tab:time_consumption}
\end{table}

\subsection{Cross-model Transferability}
LIIF was chosen as it represents foundational work in the field, while LTE and CiaoSR were selected due to their status as classical methods frequently cited in many ASSR papers, with open-source implementations readily available. To make a more comprehensive analysis, we have expanded our evaluation to include the cross-model transferability of our attack method on two additional arbitrary-scale super-resolution (ASSR) models: A-LIIF \cite{li2022adaptive} and LMF \cite{he2024latent}. Note here we do not conduct experiments on CLIT \cite{chen2023cascaded} due to the unavailability of a pre-trained model and the huge resource consumption for training. For the five ASSR models evaluated (LIIF \cite{chen2021learning}, LTE \cite{lee2022local}, CiaoSR \cite{cao2023ciaosr}, A-LIIF \cite{li2022adaptive}, and LMF \cite{he2024latent}), while they share the common approach of leveraging image encoders to extract latent features for subsequent coordinate-based super-resolution, their implementations are based on distinct high-level ideas. To be specific, LIIF uses a simple and efficient implicit function (MLP design) to map discrete pixel representations to continuous domains. LTE extends this by introducing a Local Texture Estimator, excelling in reconstructing fine-grained textures, particularly in texture-rich regions. A-LIIF enhances adaptive modeling by adjusting strategies based on local features, improving feature encoding for better detail recovery, particularly in edges and textures. CiaoSR incorporates dual attention mechanisms (Implicit Attention and Attention-in-Attention) to effectively model both global and local features. LMF combines latent modulated functions with neural field techniques, effectively capturing fine details and global image features.

Despite their distinct design emphases for arbitrary-scale super-resolution, as shown in Table~\ref{tab:cross_model_result}, we show the comparison of cross-model transferability between $\mathcal{L}_{final}$ and $\mathcal{L}_{SI}$ on B100 dataset. We can find that, with similar white-box attack performance on all five SR models (LIIF, LTE, CiaoSR, A-LIIF and LMF), the attack with $\mathcal{L}_{final}$ stably achieves better cross-model transferability than that with $\mathcal{L}_{SI}$. Figure~\ref{fig:cross-model} shows the cross-model SR output of $I^{adv}$ with $\mathcal{L}_{final}$. The SR output of $I^{adv}$ on the target model shows obvious degradation.

\subsection{Time consumption}\label{sec:time_consumption}
To provide clarity on the specific time cost of our attack method across various arbitrary-scale approaches and datasets, we present detailed time consumption per image processing with Nvidia A6000 in Table~\ref{tab:time_consumption}. Please note that the time consumption here pertains to the loss $\mathcal{L}_{final}$, resulting in a higher time consumption compared to that in Figure~\ref{fig:cost_time_memory}, which relates to the loss $\mathcal{L}_{SI}$.

\begin{figure*}[tb]
\centering
\includegraphics[width=\linewidth]{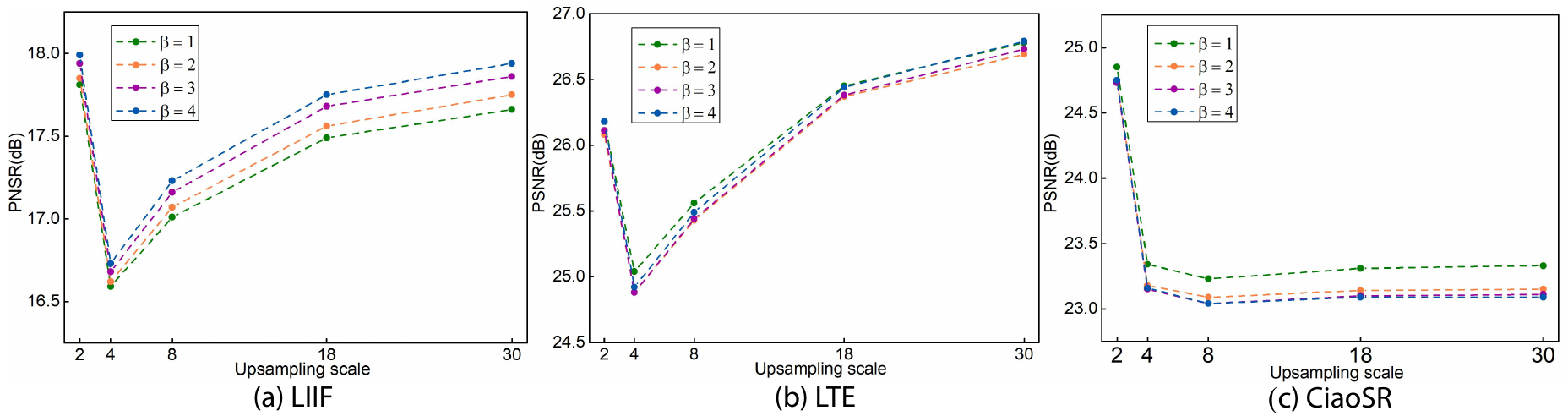}
\caption{The attack performance with different loss weights $\beta$. The adversarial images are generated by attacking LIIF.}
\label{fig:beta_ablation}
\end{figure*}

\begin{figure*}[tb]
\centering
\includegraphics[width=\linewidth]{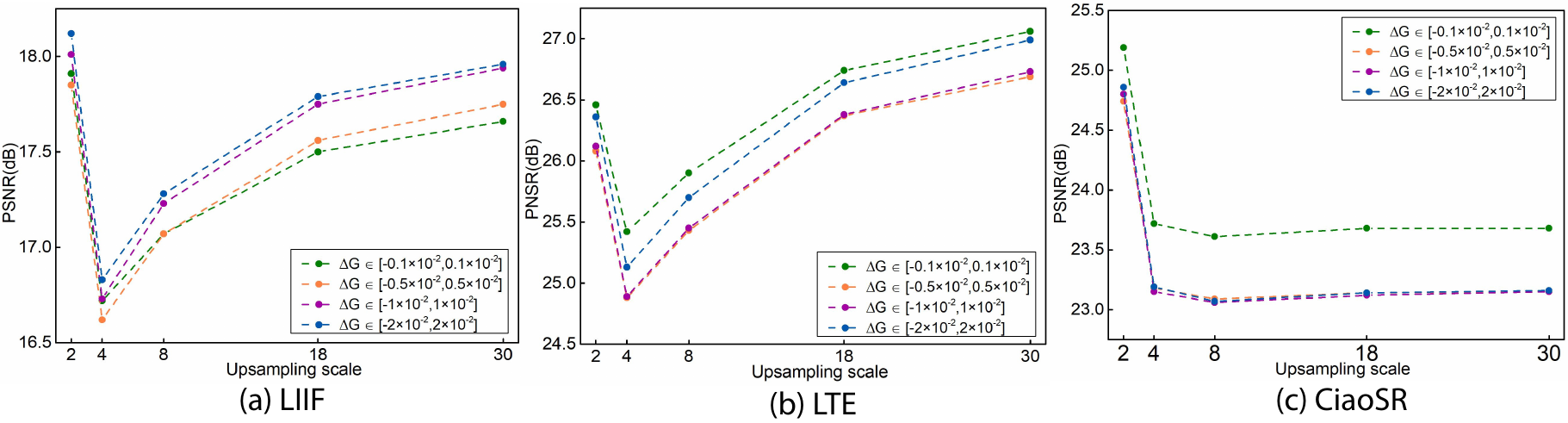}
\caption{The attack performance with different ranges of $\Delta G$. The adversarial images are generated by attacking LIIF.}
\label{fig:Delta_G_ablation}
\end{figure*}
\begin{table}[tb]
\centering
\caption{Ablation study of iteration strategy.}
\begin{tabular}{l|c|ccccc}
\hline 
\multirow{2}{*}{\makecell{Iteration \\ Magnitude $\alpha$}} & \multirow{2}{*}{LR} & \multirow{2}{*}{$\times 2$} & \multirow{2}{*}{$\times 4$} & \multirow{2}{*}{$\times 8$}& \multirow{2}{*}{$\times 18$} & \multirow{2}{*}{$\times 30$} \tabularnewline
& & & & \tabularnewline
\hline 
$\alpha = \epsilon$~/~T & 38.91 & 24.21 & 22.68 & 22.87 & 23.21 & 23.35 \tabularnewline
$\alpha = \epsilon$     & 37.69 & 26.22 & 25.31 & 25.22 & 25.27 & 25.29 \tabularnewline
$\alpha = \epsilon$~/~t & 38.36 & \textbf{17.85} & \textbf{16.62} & \textbf{17.07} & \textbf{17.56} & \textbf{17.75} \tabularnewline
\hline 
\end{tabular}
\label{tab:iteration_ablation}
\end{table}
\begin{table}[tb]
\centering
\caption{Ablation study of coordinate number per block.}
\resizebox{\linewidth}{!}{
\begin{tabular}{l|c|cccccc}
\hline 
\multirow{2}{*}{\makecell{Coordinate \\ Number $n$}} & \multirow{2}{*}{LR} & \multirow{2}{*}{$\times 2$} & \multirow{2}{*}{$\times 4$} & \multirow{2}{*}{$\times 8$} & \multirow{2}{*}{$\times 18$} & \multirow{2}{*}{$\times 30$} & \multirow{2}{*}{Memory (GB)}\tabularnewline
 &  &  &  &  &  & &\tabularnewline
\hline 
\textit{n} = 1 & 38.35 & 18.38 & 17.07 & 17.52 & 18.02 & 18.21 & 3.41\tabularnewline
\textit{n} = 2 & 38.33 & 18.06 & 16.80 & 17.26 & 17.76 & 17.94 & 3.75\tabularnewline
\textit{n} = 4 & 38.36 & 17.85 & 16.62 & 17.07 & 17.56 & 17.75 & 4.46\tabularnewline
\textit{n} = 8 & 38.32 & 17.76 & 16.55 & 17.01 & 17.50 & 17.66 & 6.65\tabularnewline
\textit{n} = 16 & 38.31 & 17.74 & 16.52 & 16.98 & 17.47 & 17.66 & 10.55\tabularnewline
\hline 
\end{tabular}
}
\label{tab:querynumber_comparision}
\end{table}

\subsection{Ablation Study}\label{sec:ablation_study}
In this section, we conduct ablation studies on the effectiveness of each component in the SIAGT method. We select LIIF \cite{chen2021learning} and B100 \cite{martin2001database} for the experiment since LIIF is the classical one based on continuous representation and B100 has a rich diversity in image content. The upsampling scale $s \in \{2,4,8,18,30\}$. 

\noindent\textbf{Iteration strategy.}
We compare our strategy with two other alternative methods. One is setting the perturbation to be $\epsilon/\mathrm{T}$ at each iteration (standard choice of PGD), and the other is setting perturbation to be $\epsilon$ at each iteration. These two strategies use fixed magnitude in each iteration and are always smaller or bigger than the iteration magnitude of our strategy. Table~\ref{tab:iteration_ablation} reports the PSNR results of using three strategies on SIAGT, and we can find that our strategy is significantly better than the others.

\noindent\textbf{Query number of coordinates per block ($n$).}
In Table~\ref{tab:querynumber_comparision}, we evaluate the influence of the query number of coordinates per block on attack performance. We select five coordinate numbers ($n \geq 1, 2, 4, 8, 16$). We can find that the attack performance is good even with $n=1$ and becomes better when the $n$ becomes bigger. However, the increasing rate slows down when $n \geq 4$ and the consumption of video memory grows dramatically. Thus, to keep the balance between attack performance and memory consumption (also with time consumption), we select $n=4$.

\begin{table*}[htb]
\centering
\caption{Adversarial LR/SR pixel adjacency differences and high-frequency information.}
\resizebox{\linewidth}{!}{
\begin{tabular}{l|c|c|c|c|c|c|c|c|c|c|c|c}
\hline 
\multirow{2}{*}{Datasets} & \multicolumn{3}{c|}{LR} & \multicolumn{3}{c|}{High-frequency Ratio (\%)} & \multicolumn{2}{c|}{$\times$2 (LTE)} & \multicolumn{2}{c|}{$\times$4 (LTE)} & \multicolumn{2}{c}{$\times$6 (LTE)} \tabularnewline
 & \multicolumn{1}{c}{clean} & \multicolumn{1}{c}{$I^{adv}$} & $I^{adv}_{TR}$ & \multicolumn{1}{c}{clean} & \multicolumn{1}{c}{$I^{adv}$} & $I^{adv}_{TR}$ & \multicolumn{1}{c}{$I^{B}$} & $I^{B}_{TR}$ & \multicolumn{1}{c}{$I^{B}$} & $I^{B}_{TR}$ & \multicolumn{1}{c}{$I^{B}$} & $I^{B}_{TR}$\tabularnewline
\hline 
Set5 & 0.1390 & 0.1467 & \textbf{0.1469} & 16.56 & 18.64 & \textbf{18.97} & 0.0669 & \textbf{0.0690} & 0.0401 & \textbf{0.0427} & 0.0266 & \textbf{0.0279}\tabularnewline
\hline 
Set14 & 0.1231 & 0.1311 & \textbf{0.1313} & 37.26 & 39.06 & \textbf{39.39} & 0.0671 & \textbf{0.0706} & 0.0411 & \textbf{0.0443} & 0.0271 & \textbf{0.0294}\tabularnewline
\hline 
B100 & 0.1047 & 0.1135 & \textbf{0.1138} & 32.03 & 34.63 & \textbf{34.96} & 0.0673 & \textbf{0.0707} & 0.0417 & \textbf{0.0448} & 0.0279 & \textbf{0.0297}\tabularnewline
\hline 
Urban100 & 0.1647 & 0.1721 & \textbf{0.1752} & 40.75 & 42.18 & \textbf{42.52} & 0.0899 & \textbf{0.0968} & 0.0552 & \textbf{0.0619} & 0.0369 & \textbf{0.0411}\tabularnewline
\hline 
\end{tabular}
}
\label{tab:pixel_difference}
\end{table*}
\begin{figure}[tb]
\centering
\includegraphics[width=\linewidth]{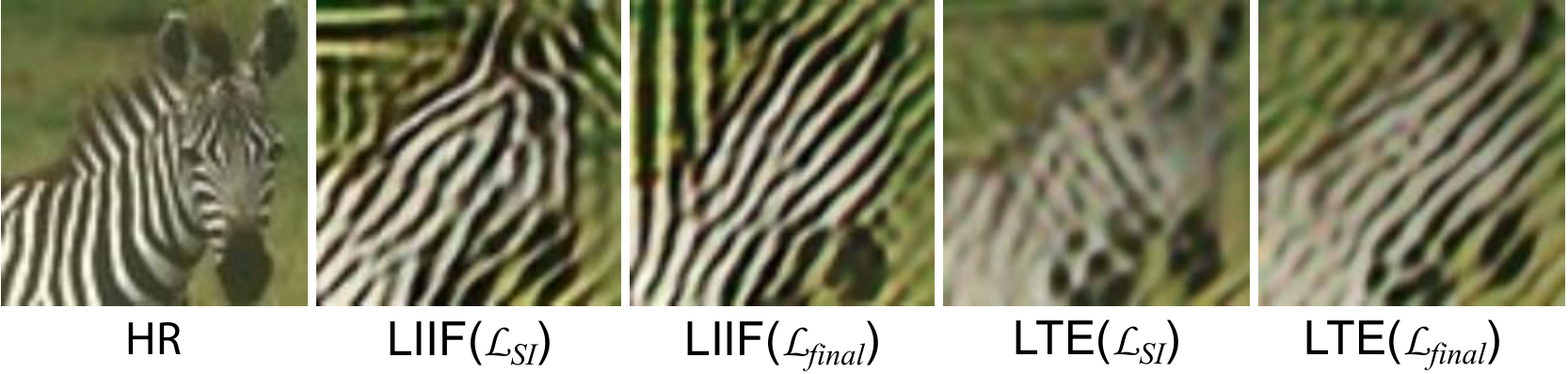}
\caption{Visual comparison of attack results with different loss functions on LIIF (source model) and cross-model testing on LTE (target model). 
SR results of $I^{adv}$ using $\mathcal{L}_{SI}$ exhibit continuity (horse outline visible), while those generated by $\mathcal{L}_{final}$ are broken on continuity.}
\label{fig:loss_ablation_visualization}
\end{figure}
\begin{table*}[tb]
\centering
\caption{Attack performance with different block numbers on Set5, Set14, B100 and Urban100 datasets.}
\begin{tabular}{l|cccccc|cccccc}
\hline 
\multirow{2}{*}{Block Number} & \multicolumn{6}{c|}{Set5} & \multicolumn{6}{c}{Set14}\tabularnewline
 & LR & $\times 2$ & $\times 4$  & $\times 8$  & $\times 18$ & $\times 30$  & LR & $\times 2$ & $\times 4$  & $\times 8$  & $\times 18$ & $\times 30$ \tabularnewline
\hline 
$H \times W$ & 37.86 & 20.94 & 19.57 & 19.97 & 20.44 & 20.62 & 38.21 & 19.31 & 18.01 & 18.47 & 18.97 & 19.15 \tabularnewline
$H~/~2 \times W~/~2$  & 38.00 & 22.36 & 20.90 & 21.28 & 21.75 & 21.92 & 38.30 & 20.65 & 19.18 & 19.62 & 20.12 & 20.31 \tabularnewline
$H~/~4  \times W~/~4$ & 38.22 & 24.71 & 23.01 & 23.31 & 23.74 & 23.90 & 38.48 & 22.58 & 20.89 & 21.28 & 21.76 & 21.94 \tabularnewline
$H~/~8  \times W~/~8$  & 38.41 & 26.06 & 24.32 & 24.62 & 25.05 & 25.21 & 38.65 & 24.31 & 22.52 & 22.85 & 23.30 & 23.48 \tabularnewline
$H~/~16  \times W~/~16$ & 38.49 & 27.39 & 25.71 & 26.00 & 26.43 & 26.59 & 38.70 & 25.94 & 24.10 & 24.34 & 24.75 & 24.91\tabularnewline
$H~/~32  \times W~/~32$ & 38.32 & 30.08 & 28.54 & 28.64 & 28.91 & 29.01 & 38.49 & 28.73 & 27.00 & 27.05 & 27.32 & 27.43 \tabularnewline

\hline 
\multirow{2}{*}{Block Number} &  \multicolumn{6}{c|}{B100} & \multicolumn{6}{c}{Urban100}\tabularnewline
& LR & $\times 2$ & $\times 4$  & $\times 8$  & $\times 18$ & $\times 30$  & LR &$\times 2$ & $\times 4$  & $\times 8$  & $\times 18$ & $\times 30$  \tabularnewline
\hline 
$H \times W$  & 38.35 & 18.38 & 17.07 & 17.52 & 18.02 & 18.21 & 38.15 & 17.91 & 16.59 & 17.10 & 17.62 & 17.81\tabularnewline
$H~/~2 \times W~/~2$ & 38.44 & 19.95 & 18.42 & 18.85 & 19.34 & 19.53& 38.26 & 19.17 & 17.69 & 18.20 & 18.73 & 18.93\tabularnewline
$H~/~4  \times W~/~4$ & 38.63 & 22.09 & 20.32 & 20.68 & 21.15 & 21.33 & 38.44 & 20.85 & 19.16 & 19.63 & 20.16 & 20.35\tabularnewline
$H~/~8  \times W~/~8$   &38.80 & 23.80 & 21.92 & 22.22 & 22.66 & 22.83 & 38.62 & 22.44 & 20.58 & 21.00 & 21.50 & 21.69\tabularnewline
$H~/~16  \times W~/~16$ &38.86 & 25.32 & 23.41 & 23.63 & 24.02 & 24.17 & 38.71 & 24.13 & 22.17 & 22.49 & 22.95 & 23.12\tabularnewline
$H~/~32  \times W~/~32$ &38.54 & 28.33 & 26.57 & 26.59 & 26.83 & 26.93 & 38.53 & 26.44 & 24.48 & 24.63 & 24.97 & 25.11\tabularnewline
\hline 
\end{tabular}
\label{tab:block_comparision}
\end{table*}

\noindent\textbf{Hyperparameter selection.} We investigate the effect of the hyperparameters utilized in the SIAGT method by applying the white-box attack on LIIF and use generated $I^{adv}$ to test cross-model transferability on LTE and CiaoSR, including the parameter $\beta$ of the $\mathcal{L}_{TR}$ loss and the range of coordinate offset $\Delta G$. Figure~\ref{fig:beta_ablation} demonstrates that when $\beta \geq 2$, the cross-model transferability is very close while the white-box attack performance gradually declines, so we select $\beta = 2$. Figure~\ref{fig:Delta_G_ablation} illustrates that the white-box attack performance decreases when the range is large, while the cross-model transferability decreases when the range is excessively small. Thus we select the range of $\Delta G$ as $[-0.5 \times 10^{-2}, 0.5 \times 10^{-2}]$ as trade-off between white-box attack performance and cross-model transferability.

\noindent\textbf{Effect of $\mathcal{L}_{TR}$.}
Here we explore the effect of $\mathcal{L}_{TR}$ in detail. Assume attacking SR model LIIF with $\mathcal{L}_{SI}$ and $\mathcal{L}_{final}$ and get deteriorated SR images $I^{A}$ and $I^{A}_{TR}$ respectively. $\mathcal{L}_{TR}$ is directly designed to result in a bigger adjacent pixel difference in $I^{A}_{TR}$ than $I^{A}$. That results in a bigger adjacent pixel difference in adversarial LR image $I^{adv}_{TR}$ than $I^{adv}$. In Table~\ref{tab:pixel_difference}, we measure the mean pixel difference between each pixel and its eight neighbors across four datasets and the intensity of high-frequency information. We can find our loss design effectively amplifies adjacent pixel differences in adversarial low-resolution images. Using Fourier transform, we converted the clean low-resolution image, the adversarial low-resolution image ($I^{adv}$) generated with only the adversarial loss $\mathcal{L}_{SI}$, and the adversarial low-resolution image ($I^{adv}_{TR}$) generated with $\mathcal{L}_{final}$ into frequency spectrums. To achieve high-frequency components, we applied the same filter radius threshold across all three images in the frequency domain. The ``High-frequency Ratio'' column reveals that the clean image consistently has the lowest ratio, while $I^{adv}$ and $I^{adv}_{TR}$ exhibit higher values across all four datasets. This contributes to the lower quality of super-resolved images of $I^{adv}$ and $I^{adv}_{TR}$, as high-frequency content is inherently challenging to restore. Furthermore, the ratio for $I^{adv}_{TR}$ is slightly but consistently higher than that of $I^{adv}$ across all datasets, reflecting a correlation between higher adjacent pixel differences and an increased high-frequency ratio. We use the SR model LTE with $I^{adv}_{TR}$ and $I^{adv}$ as input, upsampling the images with scales 2, 4, and 6. In Table~\ref{tab:pixel_difference}, there is a bigger adjacent pixel difference in output deteriorated SR images $I^{B}_{TR}$ than $I^{B}$ under different scales. $I^{B}_{TR}$ is more likely to have worse image quality than $I^{B}$. This reflects the impact of $\mathcal{L}_{TR}$ in enhancing attack transferability. The visualization difference is shown in Figure~\ref{fig:loss_ablation_visualization}.

\begin{table}[tb]
\centering
\caption{Discussion with coupling coordinate number per block and different block numbers together, with a fixed upsampling scale of 4. The experiment was performed using LIIF on the B100 dataset.}
\begin{tabular}{c|c|ccc}
\hline 
\multicolumn{2}{c|}{\multirow{2}{*}{Upsampling Scale = 4}} & \multicolumn{3}{c}{Block Number}\tabularnewline
\cline{3-5}
\multicolumn{2}{c|}{} & 6 & 24 & 96\tabularnewline
\hline 
\multirow{3}{*}{Coordinate Number $n$} & $n$ = 1 & 26.60 & 23.41 & 21.94\tabularnewline
 & $n$ = 2 & 25.12 & 22.62 & 21.14\tabularnewline
 & $n$ = 4 & 23.85 & 21.91 & 20.30\tabularnewline
\hline 
\end{tabular}
\label{tab:block_number_and_query_per_number}
\end{table}

\begin{table}[tb]
\centering
\caption{Experiment with the relationship between distortion and upsampling scale. The experiment was performed using LIIF on the B100 dataset.}
\begin{tabular}{c|ccccc}
\hline 
\multirow{2}{*}{LR} & \multicolumn{5}{c}{Upsampling Scales}\tabularnewline
\cline{2-6}
 &  $\times$1.01 &  $\times$1.05 &   $\times$1.1 &  $\times$1.5 & 
 $\times$2.0\tabularnewline
\hline 
38.36 & 27.10 & 25.87 & 24.47 & 19.15 & 17.28\tabularnewline
\hline 
\end{tabular}
\label{tab:distortion_upsampling_scale}
\end{table}

\subsection{Discussion}\label{sec:discussion}
We further explore more factors that may influence the attack performance of SIAGT on the arbitrary-scale SR. 

\noindent\textbf{Block number.} In our experiment, we empirically split the continuous image $I$ into $H \times W$ blocks. The parameters $H$ and $W$ are around 100 ($H$=84, $W$=78 in Set5; $H$=114, $W$=122 in Set14; $H$=89, $W$=110 in B100; $H$=99, $W$=123 in Ubran100), which are reasonable as the resolution of the LR image and are commonly used in arbitrary-scale SR tasks. However, to evaluate the influence of the block number on attack performance, we further exponentially reduce the block number. In Table~\ref{tab:block_comparision}, we split the continuous image $I$ into six different block numbers with one query coordinate per block (\ie, $n \geq 1$) for the attack on four datasets. We can find that the attack performance also becomes bad with the decrease in block number. However, the attack performance does not reduce much, and PSNR remains under 30 even under the situation block number is extremely small ($H$~/~32 $\times$ $W$~/~32) (\eg, 2 $\times$ 3 in B100). The result reflects the practicality and rationality of our method.

\noindent\textbf{Coupling block number and query number of coordinates per block.} In order to evaluate the effect of jointly adjusting block number (Block Number) and query number of coordinates per block (Coordinate Number $n$), as shown in Table~\ref{tab:block_number_and_query_per_number}, we fix the upsampling scale and perform an experiment using LIIF on the B100 dataset. For Coordinate Number $n$, we select 1, 2, 4. For Block Number, we split the continuous image $I$ into $H / a \times W / a$ blocks, with $a$ selected with 32, 16, and 8. When $a$ = 32, 16, 8, the corresponding block numbers in the B100 dataset are 6, 24, and 96, respectively. We observe that as the coordinate number and block number increase, greater distortion (indicated by lower PSNR values) is introduced, reflecting the interconnected effect between these two factors.

\begin{table}[tb]
\centering
\caption{Attack performance with different attack strengths.}
\resizebox{\linewidth}{!}{
\begin{tabular}{c|c|cc|cc|ccccc}
\hline 
\multirow{2}{*}{\makecell{Attack \\ Strength $\epsilon$ }} & \multirow{2}{*}{LR} & \multicolumn{2}{c|}{JND} & \multicolumn{2}{c|}{High-frequency Ratio (\%)} & \multicolumn{5}{c}{SR}\tabularnewline
 &  & $I^{adv}$ & $I^{adv}_{TR}$ & $I^{adv}$ & $I^{adv}_{TR}$ & $\times2$ & $\times4$ & $\times8$ & $\times18$ & $\times30$\tabularnewline
\hline 
1/255  & 54.75 & 1.13 & 1.12 & 32.19 & 32.31 & 37.16 & 35.77 & 35.64 & 35.74 & 35.79 \tabularnewline
2/255  & 49.15 & 1.38 & 1.38 & 32.45 & 32.58 & 28.97 & 27.14 & 27.26 & 27.60 & 27.74 \tabularnewline
4/255  & 43.77 & 2.20 & 2.19 & 33.10 & 33.29 & 21.96 & 20.36 & 20.68 & 21.12 & 21.29 \tabularnewline
8/255  & 38.36 & 3.81 & 3.79 & 34.63 & 34.96 & 17.85 & 16.62 & 17.07 & 17.56 & 17.75 \tabularnewline
16/255 & 32.52 & 7.12 & 7.07 & 38.15 & 38.58 & 15.54 & 14.63 & 15.19 & 15.72 & 15.91 \tabularnewline
\hline 
\end{tabular}
}
\label{tab:attack_strength_comparision}
\end{table}

\begin{figure}[tb]
\centering
\includegraphics[width=0.8\linewidth]{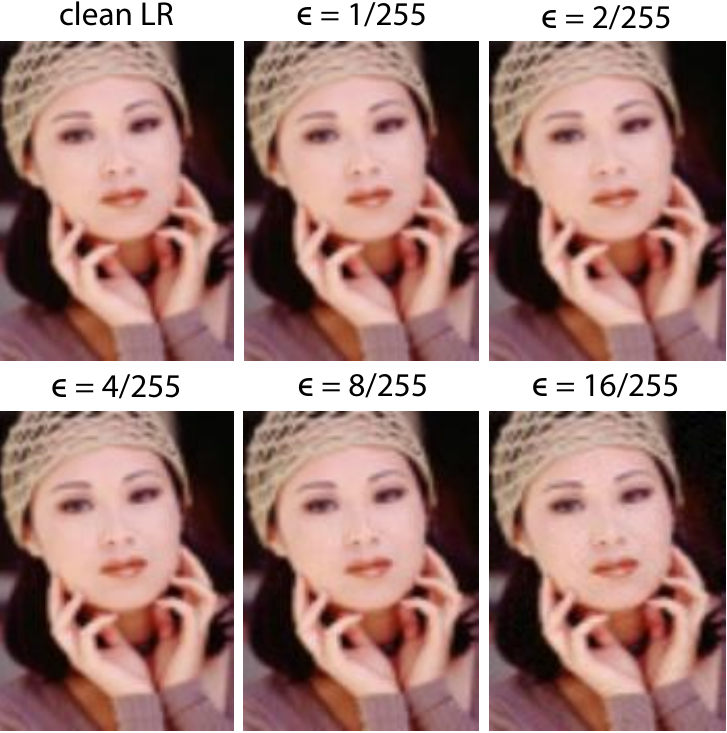}
\caption{Visualization of adversarial low-resolution image under different attack strengths.}
\label{fig:attack_strength_visualization_2}
\end{figure}

\noindent\textbf{Relationship between distortion and upsampling scale.} It is interesting to evaluate the relationship between distortion and upsampling scales. As shown in Table~\ref{tab:distortion_upsampling_scale}, we have conducted an experiment using LIIF on the B100 dataset with \textbf{different small upsampling scales}. In the LR column, it shows the PSNR between the original clean low-resolution image $I^{LR}$ and the adversarial low-resolution $I^{adv}$, while other columns (upsampling scale) show the PSNR between their corresponding SR images. We observe that distortions emerge rapidly. Specifically, when the upsampling scale is significantly small (1.01), the image dimensions only expand by a few pixels in height and width, yet the PSNR value (27.10) already falls below 30. As the upsampling scale increases slightly (to 1.5), the PSNR value (19.15) drops further to below 20.

\noindent\textbf{Attack strength.}
In Table~\ref{tab:attack_strength_comparision}, we explore the imperceptibility and effectiveness of our method under different attack strengths. With the increase of attack strength, the attack performance on different upsampling scales all becomes better, and the PSNR of SR images becomes close to or less than 20 when the attack strength is bigger than 4/255. With regard to the imperceptibility of the attack, we propose that the stealthiness of the attack is determined entirely by $\epsilon$, not by the loss function design. As shown in Table~\ref{tab:attack_strength_comparision}, we evaluate the high-frequency ratio of adversarial low-resolution images, and the corresponding just noticeable difference (JND) \cite{wikipedia_color_difference, wikipedia_jnd} between adversarial low-resolution images and clean images. To be specific, we calculate the JND for both the adversarial low-resolution image ($I^{adv}$) generated using only the adversarial loss $\mathcal{L}_{SI}$ and the adversarial low-resolution image ($I^{adv}_{TR}$) generated using $\mathcal{L}_{final}$. As shown in the ``JND'' column of Table~\ref{tab:attack_strength_comparision}, the JND value (where higher values indicate greater detectability) increases with the attack strength $\epsilon$, demonstrating the impact of attack strength on the imperceptibility of the attack. We also calculate the high-frequency ratio. Using the Fourier transform, $I^{adv}$ and $I^{adv}_{TR}$ are converted into frequency spectrums, and a consistent filter radius threshold is applied in the frequency domain to isolate high-frequency components. The ``High-frequency Ratio'' column shows that the ratio for $I^{adv}_{TR}$ is slightly but consistently higher than that of $I^{adv}$ across all five attack strengths. However, the corresponding JND values of $I^{adv}_{TR}$ are slightly lower or equal to those of $I^{adv}$. This indicates that increased high-frequency information does not lead to higher JND values, suggesting that it is basically unrelated to the imperceptibility of the attack. Consequently, this suggests that the design of the $I^{adv}_{TR}$ loss function has minimal impact on the attack's imperceptibility. We show the adversarial low-resolution images under different attack strengths in Figure~\ref{fig:attack_strength_visualization_2}.

\noindent\textbf{Compare with alternative methods.}
Here we compare SIAGT with three alternative methods. (1) Although the baseline \cite{choi2019evaluating} is time-consuming and memory-consuming, we further compare with it to see the gap between attack performance. (2) Just as mentioned in Section~\ref{sec:problem_formulation_motivation}, discarding coordinates and only leveraging the the latent code $\mathcal{Z}$ to construct the loss $\mathcal{L}_{latent}$ is an alternative method. (3) We further show the effect of only using $\mathcal{L}_{SI}$. As shown in Table~\ref{tab:baseline_comparision},
we show the comparison on five upsampling scales (for the baseline, a bigger upsampling scale will lead to memory overflow) across four datasets. We can find that baseline only achieves a bit better attack performance than SIAGT, which means the attack performance of SIAGT is enough with the advantage of resource-saving. 
Meanwhile, the attack performance of SIAGT is similar to the attack with $\mathcal{L}_{SI}$ and both are much better than the attack with $\mathcal{L}_{latent}$, which fully shows the necessity and effectiveness of our method.

\begin{table}[tb]
\centering
\caption{Comparing SIAGT with baseline and alternative methods on four datasets under different upsampling scales.}
\begin{tabular}{l|l|ccccc}
\hline 
Datasets & Method & $ \times 2$ & $ \times 3$ & $ \times 4$ & $ \times 6$ & $ \times 8$\tabularnewline
\hline 
\multirow{4}{*}{Set5} & SIAGT & 20.34 & 19.07 & 19.05 & 19.24 & 19.41\tabularnewline
 & Baseline \cite{choi2019evaluating}& 20.30 & 19.04 & 18.84 & 18.71 & 19.10\tabularnewline
 & $\mathcal{L}_{latent}$ & 22.05 & 20.87 & 20.90 & 20.92 & 21.00\tabularnewline
 & $\mathcal{L}_{SI}$ & 20.33 & 19.11 & 19.07 & 19.21 & 19.38\tabularnewline
\hline 
\multirow{4}{*}{Set14} & SIAGT & 18.70 & 17.44 & 17.50 & 17.74 & 17.98\tabularnewline
 & Baseline & 18.51 & 17.15 & 17.19 & 17.31 & N/A\tabularnewline
 & $\mathcal{L}_{latent}$ & 21.45 & 20.33 & 20.36 & 20.35 & 20.40\tabularnewline
 & $\mathcal{L}_{SI}$ & 18.65 & 17.50 & 17.50 & 17.65 & 17.83\tabularnewline
\hline 
\multirow{4}{*}{B100} & SIAGT & 17.85 & 16.56 & 16.62 & 16.85 & 17.07\tabularnewline
 & Baseline & 17.76 & 16.42 & 16.37 & 16.44 & 16.55\tabularnewline
 & $\mathcal{L}_{latent}$ & 20.24 & 19.13 & 19.17 & 19.15 & 19.20\tabularnewline
 & $\mathcal{L}_{SI}$ & 17.91 & 16.67 & 16.70 & 16.86 & 17.08\tabularnewline
\hline 
\multirow{4}{*}{Urban100} & SIAGT & 17.41 & 16.10 & 16.15 & 16.42 & 16.66\tabularnewline
 & Baseline & 16.85 & 15.66 & 15.59 & 15.66 & 15.87\tabularnewline
 & $\mathcal{L}_{latent}$ & 19.90 & 18.69 & 18.70 & 18.71 & 18.78\tabularnewline
 & $\mathcal{L}_{SI}$ & 17.01 & 15.85 & 15.84 & 16.00 & 16.18\tabularnewline
\hline 
\end{tabular}
\label{tab:baseline_comparision}
\end{table}
\begin{table}[tb]
\centering
\caption{SR performance of LIIF and fine-tuned LIIF.}
\resizebox{\linewidth}{!}{
\begin{tabular}{l|c|c|c|c|c|c}
\hline 
Model & $\times$2  & $\times$3  & $\times$4 & $\times$6 & $\times$8 & CLEVER $\downarrow$ \tabularnewline
\hline 
LIIF & 32.1745 & 29.1005 & 27.5963 & 25.8427 & 24.7865 & 0.1645\tabularnewline
\hline 
Fine-tuned LIIF-clean\&adv & \textbf{32.1808} & \textbf{29.1077} & \textbf{27.6031} & \textbf{25.8531} & \textbf{24.7947} & \textbf{0.1225} \tabularnewline
\hline 
Fine-tuned LIIF-clean only & 32.1784 & 29.1070 & 27.6016 &25.8475&24.7917&0.1642
\tabularnewline
\hline 
\end{tabular}
}
\label{tab:sr_performance}
\end{table}

\begin{table}[t]
\centering
\caption{Comparison of cross-model transferability of different attack methods on B100 dataset.}
\begin{tabular}{l|c|ccccc}
\hline 
\multirow{2}{*}{Source Model} & \multirow{2}{*}{Method} & \multicolumn{5}{c}{LIIF (Target)}\tabularnewline
 &  & $\times 2$ & $\times 4$ & $\times 8$ & $\times 18$ & $\times 30$\tabularnewline
\hline 
\multirow{3}{*}{LIIF} & DIM \cite{xie2019improving} & 33.78 & 33.14  & 33.02 & 33.00 & 33.00\tabularnewline
 & ILA \cite{huang2019enhancing} & 20.41  & 19.34  & 19.37  & 19.53  & 19.59\tabularnewline
 & SIAGT & \textbf{17.85} & \textbf{16.62} & \textbf{17.07} & \textbf{17.56} & \textbf{17.75}\tabularnewline
\hline 
\multirow{3}{*}{LTE} & DIM & 34.88 & 34.27 & 34.13 & 34.11  & 34.10\tabularnewline
 & ILA & 32.45 & 31.54  & 31.38  & 31.36  & 31.36\tabularnewline
 & SIAGT & \textbf{26.71} & \textbf{25.17} & \textbf{25.03} & \textbf{25.19} & \textbf{25.29}\tabularnewline
\hline 
\multirow{3}{*}{CiaoSR} & DIM & 31.38 & 30.76 & 30.65 & 30.63 & 30.63\tabularnewline
 & ILA & 30.43 & 29.67 & 29.53 & 29.51 & 29.51\tabularnewline
 & SIAGT & \textbf{28.05} & \textbf{27.01} & \textbf{26.82} & \textbf{26.81} & \textbf{26.83}\tabularnewline
 \hline
\end{tabular}
\label{tab:input_trans}
\end{table}

\noindent\textbf{Enhancing performance of SR models.}
Adversarial attacks reveal weaknesses of SR and, similar to using adversarial examples for recognition enhancement \cite{xie2020adversarial}, using adversarial examples as data augmentation \cite{jia2024fast,li2023learning,li2024safeear} can improve the performance of the SR model. 
In Table~\ref{tab:sr_performance}, we fine-tune the LIIF with 8,000 clean images and 100 adversarial examples (third row), resulting in a fine-tuned LIIF. We use PSNR to evaluate the image quality and CLEVER \cite{choi2019evaluating} as a robustness measure (a metric independent of attack methods). The fine-tuned LIIF achieves higher PSNR on clean images and better robustness than the original LIIF, highlighting the positive effects of adversarial attacks on improving SR performance, thus demonstrating the necessity of utilizing adversarial attacks to evaluate super-performance tasks. It is also better than fine-tuned LIIF with only 8,000 clean images (fourth row).

\begin{table*}[tb]
\centering
\caption{Evaluating our SIAGT method with human perception-related metric LPIPS on the B100 dataset.}
\resizebox{0.8\linewidth}{!}{
\begin{tabular}{l|c|c|ccccccccc}
\hline 
Model & Metrics & LR & $\times 2$ & $\times 3$ & $\times 4$ & $\times 6$ & $\times 8$ & $\times 12$ & $\times 18$ & $\times 24$ & $\times 30$\tabularnewline
\hline 
\multirow{2}{*}{LIIF} & PSNR $\uparrow$ & 38.36 & 17.85 & 16.56 & 16.62 & 16.85 & 17.07 & 17.37 & 17.56 & 17.69 & 17.75\tabularnewline
 & LPIPS $\downarrow$ & 0.0369 & 0.5670 & 0.5725 & 0.5776 & 0.5867 & 0.5964 & 0.6221 & 0.6304 & 0.6137 & 0.5898\tabularnewline
\hline 
\multirow{2}{*}{LTE} & PSNR & 38.49 & 18.25 & 17.12 & 17.14 & 17.73 & 18.37 & 19.26 & 19.90 & 20.23 & 20.42\tabularnewline
 & LPIPS & 0.0418 & 0.5783 & 0.5814 & 0.5852 & 0.5876 & 0.5903 & 0.6033 & 0.6068 & 0.5883 & 0.5625\tabularnewline
\hline 
\multirow{2}{*}{CiaoSR} & PSNR & 32.80 & 12.97 & 13.11 & 12.35 & 12.56 & 12.58 & 12.73 & 12.78 & 12.84 & 12.89\tabularnewline
 & LPIPS & 0.1183 & 0.6623 & 0.6616 & 0.6688 & 0.6822 & 0.6915 & 0.7085 & 0.7088 & 0.6887 & 0.6687\tabularnewline
\hline 
\end{tabular}
}
\label{tab:More_perception_metrics}
\end{table*}
\begin{table}[tb]
\centering
\caption{Comparison with other selection strategies.}
\resizebox{\linewidth}{!}{
\begin{tabular}{l|c|ccccc}
\hline 
Strategies & LR & $\times 2$ & $\times 4$ & $\times 8$ & $\times 18$ & $\times 30$\tabularnewline
\hline 
High-frequency & 38.44 & 20.36 & 18.96 & 19.29 & 19.70 & 19.85\tabularnewline
\hline 
Low-frequency & 38.34 & 18.06 & 16.84 & 17.19 & 17.62 & 17.78\tabularnewline
\hline 
Random Selection & 38.34 & \textbf{17.91} & \textbf{16.70} & \textbf{17.08} & \textbf{17.47} & \textbf{17.63}\tabularnewline
\hline  
\end{tabular}
}
\label{tab:select_comparision}
\end{table}

\noindent\textbf{Cross-model transferability of methods from classification.}
In image classification, input transformation (\eg, DIM~\cite{xie2019improving}) and intermediate feature attack (\eg, ILA~\cite{huang2019enhancing}) are classical approaches for improving the cross-model transferability of adversarial samples. An intuitive idea is to apply them directly to the attack task on super-resolution. In Table~\ref{tab:input_trans}, we can observe that DIM cannot even attack LIIF successfully and ILA does not show a good ability to improve the cross-model transferability as our method.

\noindent\textbf{Evaluation with human perception-related metric.} Since PSNR is only weakly related to human perception, to better assess the image quality of super-resolved images from a perceptual standpoint, we further explore a \textbf{widely adopted metric} \cite{xu2024uncovering,park2023perception,qiu2023sc} that aligns more closely with human visual perception than the PSNR metric: the Learned Perceptual Image Patch Similarity (LPIPS), where lower scores (ranging from 0 to 1) indicate better perceptual similarity \cite{zhang2018unreasonable}. LPIPS aligns well with human perception \cite{zhang2018unreasonable} by: \ding{182} Capturing Key Visual Features: It is sensitive to texture, structure, and fine details critical to human perception. \ding{183} Recognizing Distortions: It effectively identifies blur, color shifts, and texture loss, aligning with subjective quality. \ding{184} Proven Correlation: It closely matches human ratings, making it a reliable metric for human-centric image evaluations. In Table~\ref{tab:More_perception_metrics} we present our results, demonstrating that adversarial low-resolution images generated by our SIAGT method closely resemble the original clean low-resolution images (with high PSNR and low LPIPS). Conversely, their super-resolved versions show low similarity (with low PSNR and high LPIPS), reflecting the effectiveness of our method in degrading the perceived quality of super-resolved images under human perception-related metrics.

\noindent\textbf{Evaluation by human.} We conduct a questionnaire survey with 60 human participants to assess image quality. The evaluation consisted of two parts: The first part compares the image quality between the original clean low-resolution image $I^{LR}$ and adversarial low-resolution image $I^{adv}$. The second part compares the image quality of super-resolution (SR) images generated from them. Each part involved 10 pairs of images. For each pair, participants are asked to evaluate which image has higher quality or indicate if they could not decide. \ding{182} For the first part ($I^{LR}$ \textit{vs.} $I^{adv}$), we find participants are unable to decide in 44.49\% of cases, find the $I^{adv}$ to be better in 28.99\% and prefer $I^{LR}$ in 26.52\%. These results highlight that humans struggle to discern differences in image quality between the clean low-resolution image $I^{LR}$ and the adversarial low-resolution image $I^{adv}$, underscoring the imperceptibility of our attack. \ding{183} For the second part (SR of $I^{LR}$ \textit{vs.} SR of $I^{adv}$), we find participants are unable to decide in 3.51\% of cases, find the SR of $I^{adv}$ to be better in 12.16\% and prefer SR of $I^{LR}$ in 84.33\%. These findings emphasize that humans can readily discern the superior image quality of the SR of the clean low-resolution image $I^{LR}$ compared to the SR of the adversarial low-resolution image $I^{adv}$, demonstrating the effectiveness of our attack in degrading the quality of the SR image. Overall, in most cases, our attack remains imperceptible, while the super-resolved image generated from the adversarial low-resolution input is perceived by humans as having lower image quality.

\noindent\textbf{Query coordinate selection.}
For the selection strategy of coordinates, excluding the random selection strategy, we conduct an extra experiment by selecting coordinates in high-frequency or low-frequency areas of the image because the effect of super-resolution is frequency-sensitive. As shown in Table~\ref{tab:select_comparision}, we find that random selection achieves the best attack performance. We think this is because random selection is adaptive to images of different diversities. Also, random selection is simple and time-saving. Thus we use the random selection strategy in SIAGT.

\begin{figure}[tb]
\centering
\includegraphics[width=\linewidth]{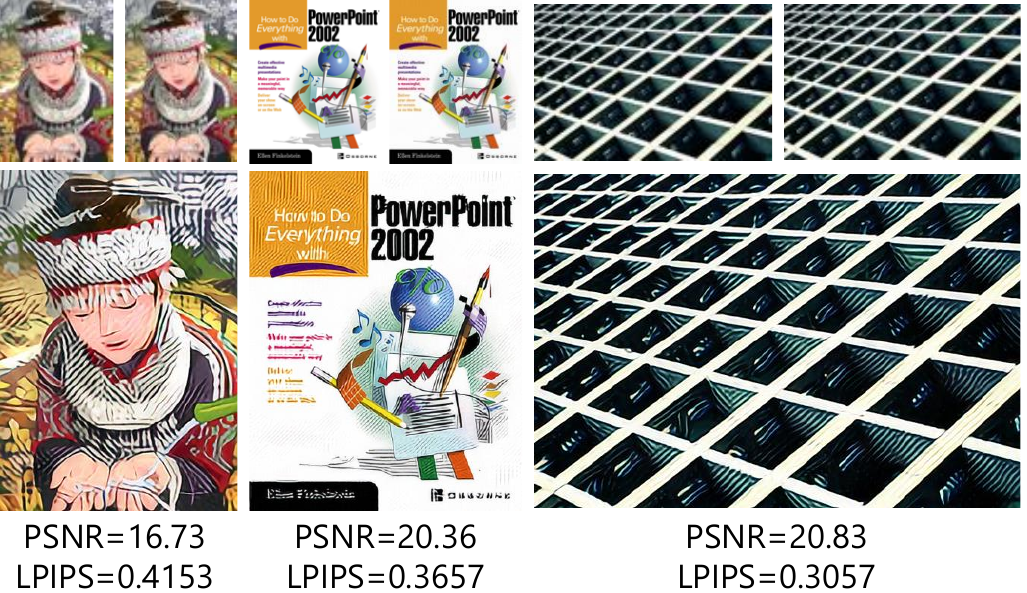}
\caption{Effect of attack on images with/without writing.}
\label{fig:has_text_quality_comparison}
\end{figure}

\noindent\textbf{Attack effect on images with/without writing.}
We conducted further evaluations on the attack's impact on images with and without writing. As shown in Figure~\ref{fig:has_text_quality_comparison}, the results are divided into three parts: two without writing and one with writing (center). Each part includes the clean low-resolution image, its adversarial counterpart, and the corresponding super-resolved (SR) image. The findings reveal that the attack on the image with writing demonstrates a moderate effect (\ie, the PSNR and LPIPS values of the SR image for the center image fall between those of the left and right images). Therefore, we suggest that the presence of writing does not significantly influence whether the attack effect is better or worse.

\noindent\textbf{Influence on downstream tasks.}
Since adversarial examples can significantly deteriorate SR images, there are potential risks when SR models are used as pre-processing modules. In order to investigate the effect of arbitrary-scale SR adversarial attacks on downstream tasks: text recognition, face detection, and street view house number recognition. 

For text recognition, we employ LIIF and compare the text recognition accuracy of original clean LR images $I^{LR}$, SR images generated from $I^{LR}$, and SR images generated from $I^{adv}$. We select TextZoom \cite{wang2020scene} as the testing dataset, which is a real scene SR dataset widely employed for text SR tasks. The dataset comprises 4,373 LR text images and is divided into easy, medium, and hard subsets based on the degree of degradation. We evaluate text images using the pre-trained text recognizer MORAN \cite{luo2019moran}. The recognition accuracy ($\uparrow$) results are shown in Table~\ref{tab:accuracy_comparision}. We can find that SR models can help to improve the recognition accuracy by 0.9\% on average while using $I^{adv}$ as input will lead to a significant reduction of recognition accuracy (17\%). 

For the face detection task, we use the classical Retinaface model \cite{deng2019retinaface} on the WIDER FACE dataset \cite{yang2016wider}. The WIDER FACE dataset is divided into Easy, Medium, and Hard subsets based on detection difficulty. We use the 4$\times$ downsampled version of the original image as the clean LR in our experiment. The face detection accuracy ($\uparrow$) results are shown in Table~\ref{tab:accuracy_face_comparision}. We can find that SR models can help to improve the detection accuracy by 20.73\% on average while using $I^{adv}$ as input will lead to a significant reduction of detection accuracy (36.28\%). 

\begin{table}[tb]
\centering
\caption{A comparison of recognition accuracy ($\uparrow$) with $I^{LR}$, SR of $I^{LR}$ and SR of $I^{adv}$ on the text recognition dataset TextZoom \cite{wang2020scene}.}
\begin{tabular}{l|cccc}
\hline 
\multirow{2}{*}{Image} & \multicolumn{4}{c}{TextZoom}\tabularnewline
 & easy & medium & hard & average\tabularnewline
\hline 
$I^{LR}$ & 58.6\% & 36.6\% & 29.3\% & 42.5\%\tabularnewline
SR of $I^{LR}$& 60.0\% & 37.6\% & 29.3\% & 43.4\%\tabularnewline
SR of $I^{adv}$ & 28.1\% & 24.7\% & 26.2\% & 26.4\%\tabularnewline
\hline 
\end{tabular}
\label{tab:accuracy_comparision}
\end{table}
\begin{table}[tb]
\centering
\caption{A comparison of face detection accuracy ($\uparrow$) with $I^{LR}$, SR of $I^{LR}$ and SR of $I^{adv}$ on the WIDER FACE dataset \cite{yang2016wider}.}
\begin{tabular}{l|cccc}
\hline 
\multirow{2}{*}{Image} & \multicolumn{4}{c}{WIDER FACE}\tabularnewline
 & easy & medium & hard & average\tabularnewline
\hline 
$I^{LR}$ & 81.59\% & 61.13\% & 26.09\% & 56.27\%\tabularnewline
SR of $I^{LR}$& 91.83\% & 85.66\% & 53.53\% & 77.00\%\tabularnewline
SR of $I^{adv}$ & 29.41\% &20.86\% &9.71\% & 19.99\%\tabularnewline
\hline 
\end{tabular}
\label{tab:accuracy_face_comparision}
\end{table}

For the street view house number recognition task, we employ Baidu's open-source Optical Character Recognition (OCR) model \cite{baidu_ocr} on the widely-used SVHN dataset \cite{netzer2011reading}. The SVHN dataset comprises real-world images of house numbers captured from Google Street View. In our experiments, we utilize the original image as the clean low-resolution (LR) input. The OCR recognition accuracy on $I^{LR}$ is 54.25\%, while the super-resolved (SR) version of $I^{LR}$ improves the accuracy to 55.15\%. In contrast, applying SR to adversarially perturbed inputs ($I^{adv}$) yields a significantly reduced recognition accuracy of 25.02\%, underscoring the adverse effects of adversarial samples on downstream tasks.

\section{Limitation}\label{sec:limitation}
Since our method is a scale-invariant attack tailored for arbitrary-scale super-resolution, the loss term, $\mathcal{L}_{SI}$, relies on a unique coordinate factor specific to this context. Consequently, it cannot be applied to fixed-scale super-resolution methods, limiting the generality of our attack.

\section{Conclusion}
In this paper, we propose a scale-invariant attack method SIAGT against arbitrary-scale SR. The high-level idea is not only constructing the attack on the continuous representation but also changing the appearance of the continuous representation by adjusting the finite discrete points in the continuous representation. 
We also raise the problem of improving the cross-model transferability of the attack on arbitrary-scale SR tasks and propose coordinate-dependent loss. In the future, we aim to construct adversarial attacks on other continuous representation-based models and tasks (\eg, Nerf \cite{mildenhall2021nerf}).

\section*{Appendix}
\section{More Visual Results}
\begin{figure*}[tb]
\centering
\includegraphics[width=0.9\linewidth]{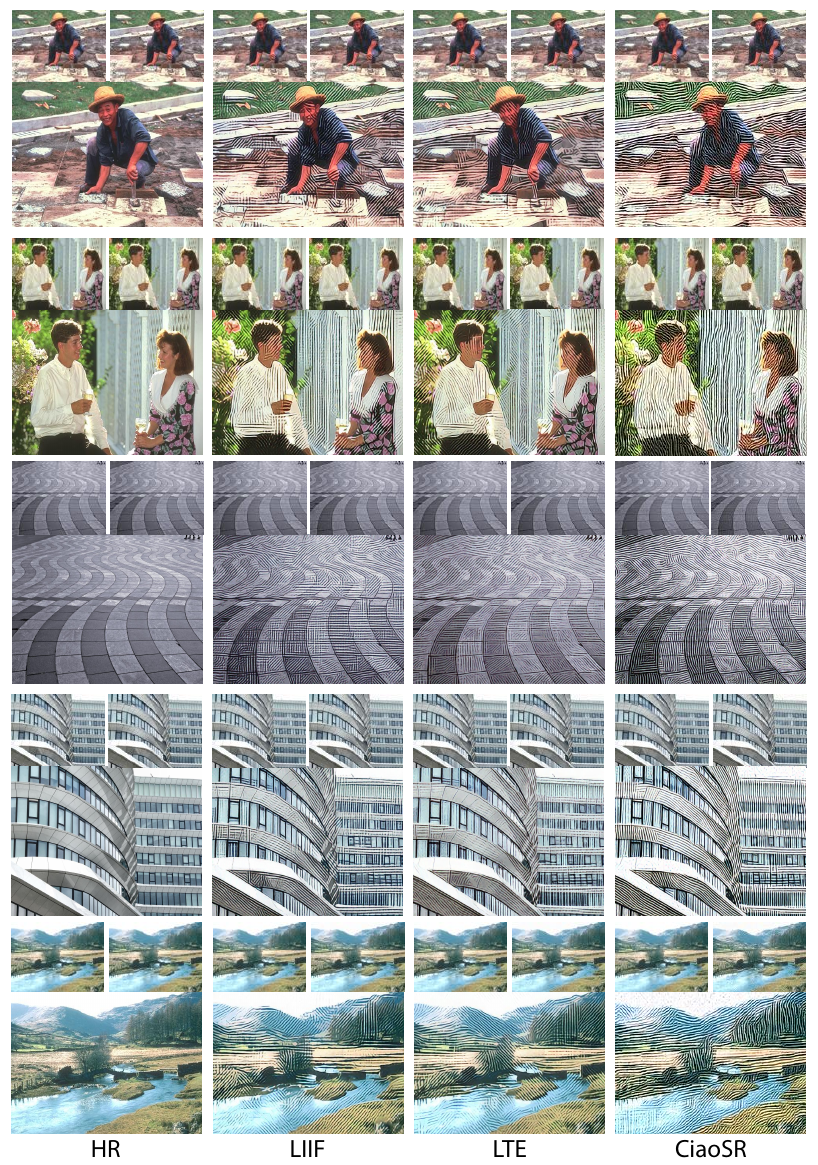}
\caption{Visualization results of attacks on different scene images. (top-left) is the original input clean image, (top-right) is the adversarial image, and (bottom) is the SR output ($\times 8$) obtained from the adversarial images.}
\label{fig:attack_result}
\end{figure*}

\begin{figure*}[tb]
\centering
\includegraphics[width=0.9\linewidth]{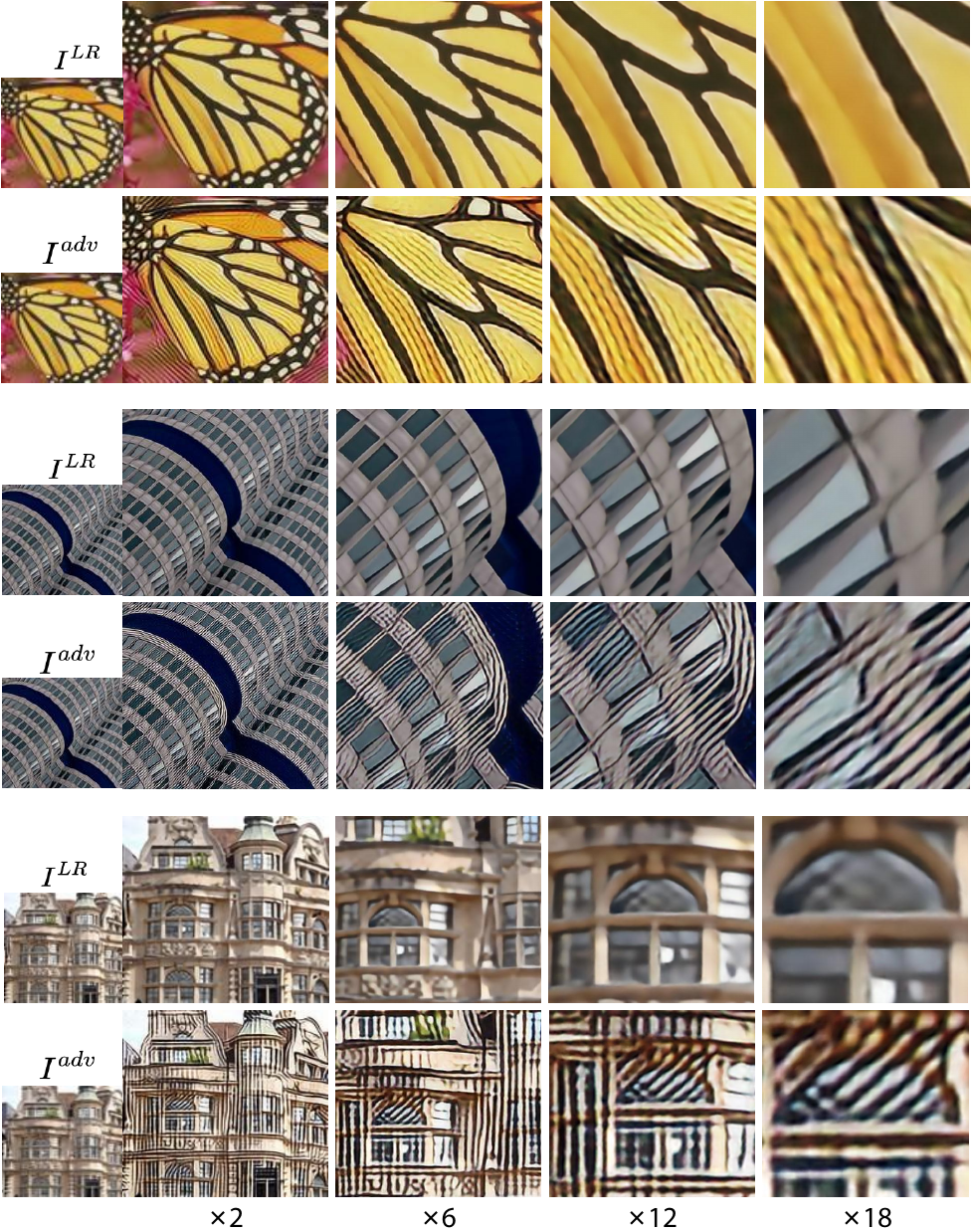}
\caption{Visualization comparison of the super-resolved results with multiple scale factors generated from clean LR image $I^{LR}$ and adversarial image $I^{adv}$.}
\label{fig:scale_vis}
\end{figure*}
In this section, we show more SR images of adversarial examples generated by the SIAGT method on datasets including Set5 \cite{bevilacqua2012low}, Set14 \cite{zeyde2012single}, B100 \cite{martin2001database}, and Urban100 \cite{huang2015single}. Specifically, we show super-resolved results from different types of adversarial LR images in Figure~\ref{fig:attack_result}, including people scene images, urban scene images, and natural scene images. Meanwhile, we also show a visualization comparison of the super-resolved results with multiple scale factors generated from clean LR image $I^{LR}$ and adversarial image $I^{adv}$ in Figure~\ref{fig:scale_vis}. It can be observed that our method shows remarkable attack performance (significantly deteriorates the SR images) across different images and upsampling scales while introducing imperceptible distortion to the targeted low-resolution (LR) images. 

\ifCLASSOPTIONcaptionsoff
  \newpage
\fi



%

\bibliographystyle{IEEEtran}
\bibliography{sample-base}

\begin{thebibliography}{10}
\providecommand{\url}[1]{#1}
\csname url@samestyle\endcsname
\providecommand{\newblock}{\relax}
\providecommand{\bibinfo}[2]{#2}
\providecommand{\BIBentrySTDinterwordspacing}{\spaceskip=0pt\relax}
\providecommand{\BIBentryALTinterwordstretchfactor}{4}
\providecommand{\BIBentryALTinterwordspacing}{\spaceskip=\fontdimen2\font plus
\BIBentryALTinterwordstretchfactor\fontdimen3\font minus \fontdimen4\font\relax}
\providecommand{\BIBforeignlanguage}[2]{{%
\expandafter\ifx\csname l@#1\endcsname\relax
\typeout{** WARNING: IEEEtran.bst: No hyphenation pattern has been}%
\typeout{** loaded for the language `#1'. Using the pattern for}%
\typeout{** the default language instead.}%
\else
\language=\csname l@#1\endcsname
\fi
#2}}
\providecommand{\BIBdecl}{\relax}
\BIBdecl

\bibitem{chen2021pre}
H.~Chen, Y.~Wang, T.~Guo, C.~Xu, Y.~Deng, Z.~Liu, S.~Ma, C.~Xu, C.~Xu, and W.~Gao, ``Pre-trained image processing transformer,'' in \emph{CVPR}, 2021, pp. 12\,299--12\,310.

\bibitem{dai2019second}
T.~Dai, J.~Cai, Y.~Zhang, S.-T. Xia, and L.~Zhang, ``Second-order attention network for single image super-resolution,'' in \emph{CVPR}, 2019, pp. 11\,065--11\,074.

\bibitem{liang2021swinir}
J.~Liang, J.~Cao, G.~Sun, K.~Zhang, L.~Van~Gool, and R.~Timofte, ``Swinir: Image restoration using swin transformer,'' in \emph{ICCV}, 2021, pp. 1833--1844.

\bibitem{mei2021image}
Y.~Mei, Y.~Fan, and Y.~Zhou, ``Image super-resolution with non-local sparse attention,'' in \emph{CVPR}, 2021, pp. 3517--3526.

\bibitem{chen2021learning}
Y.~Chen, S.~Liu, and X.~Wang, ``Learning continuous image representation with local implicit image function,'' in \emph{CVPR}, 2021, pp. 8628--8638.

\bibitem{xu2021ultrasr}
X.~Xu, Z.~Wang, and H.~Shi, ``Ultrasr: Spatial encoding is a missing key for implicit image function-based arbitrary-scale super-resolution,'' \emph{arXiv preprint arXiv:2103.12716}, 2021.

\bibitem{liu2021enhancing}
Y.-T. Liu, Y.-C. Guo, and S.-H. Zhang, ``Enhancing multi-scale implicit learning in image super-resolution with integrated positional encoding,'' \emph{arXiv preprint arXiv:2112.05756}, 2021.

\bibitem{lee2022local}
J.~Lee and K.~H. Jin, ``Local texture estimator for implicit representation function,'' in \emph{CVPR}, 2022, pp. 1929--1938.

\bibitem{cao2023ciaosr}
J.~Cao, Q.~Wang, Y.~Xian, Y.~Li, B.~Ni, Z.~Pi, K.~Zhang, Y.~Zhang, R.~Timofte, and L.~Van~Gool, ``Ciaosr: Continuous implicit attention-in-attention network for arbitrary-scale image super-resolution,'' in \emph{CVPR}, 2023, pp. 1796--1807.

\bibitem{choi2019evaluating}
J.-H. Choi, H.~Zhang, J.-H. Kim, C.-J. Hsieh, and J.-S. Lee, ``Evaluating robustness of deep image super-resolution against adversarial attacks,'' in \emph{ICCV}, 2019, pp. 303--311.

\bibitem{xun2024minimalism}
Y.~Xun, X.~Jia, J.~Gu, X.~Liu, Q.~Guo, and X.~Cao, ``Minimalism is king! high-frequency energy-based screening for data-efficient backdoor attacks,'' \emph{IEEE Transactions on Information Forensics and Security}, 2024.

\bibitem{castillo2021generalized}
A.~Castillo, M.~Escobar, J.~C. P{\'e}rez, A.~Romero, R.~Timofte, L.~Van~Gool, and P.~Arbelaez, ``Generalized real-world super-resolution through adversarial robustness,'' in \emph{Proceedings of the IEEE/CVF International Conference on Computer Vision}, 2021, pp. 1855--1865.

\bibitem{aakerberg2024pda}
A.~Aakerberg, M.~El~Helou, K.~Nasrollahi, and T.~Moeslund, ``Pda-rwsr: Pixel-wise degradation adaptive real-world super-resolution,'' in \emph{Proceedings of the IEEE/CVF Winter Conference on Applications of Computer Vision}, 2024, pp. 4097--4107.

\bibitem{goodfellow2014explaining}
I.~J. Goodfellow, J.~Shlens, and C.~Szegedy, ``Explaining and harnessing adversarial examples,'' \emph{arXiv preprint arXiv:1412.6572}, 2014.

\bibitem{kurakin2016adversarial}
A.~Kurakin, I.~Goodfellow, and S.~Bengio, ``Adversarial machine learning at scale,'' \emph{arXiv preprint arXiv:1611.01236}, 2016.

\bibitem{madry2017towards}
A.~Madry, A.~Makelov, L.~Schmidt, D.~Tsipras, and A.~Vladu, ``Towards deep learning models resistant to adversarial attacks,'' \emph{arXiv preprint arXiv:1706.06083}, 2017.

\bibitem{feng2023robust}
W.~Feng, N.~Xu, T.~Zhang, B.~Wu, and Y.~Zhang, ``Robust and generalized physical adversarial attacks via meta-gan,'' \emph{IEEE Transactions on Information Forensics and Security}, 2023.

\bibitem{ledig2017photo}
C.~Ledig, L.~Theis, F.~Husz{\'a}r, J.~Caballero, A.~Cunningham, A.~Acosta, A.~Aitken, A.~Tejani, J.~Totz, Z.~Wang \emph{et~al.}, ``Photo-realistic single image super-resolution using a generative adversarial network,'' in \emph{CVPR}, 2017, pp. 4681--4690.

\bibitem{lim2017enhanced}
B.~Lim, S.~Son, H.~Kim, S.~Nah, and K.~Mu~Lee, ``Enhanced deep residual networks for single image super-resolution,'' in \emph{CVPR workshops}, 2017, pp. 136--144.

\bibitem{zhang2018residual}
Y.~Zhang, Y.~Tian, Y.~Kong, B.~Zhong, and Y.~Fu, ``Residual dense network for image super-resolution,'' in \emph{CVPR}, 2018, pp. 2472--2481.

\bibitem{lai2017deep}
W.-S. Lai, J.-B. Huang, N.~Ahuja, and M.-H. Yang, ``Deep laplacian pyramid networks for fast and accurate super-resolution,'' in \emph{CVPR}, 2017, pp. 624--632.

\bibitem{mildenhall2021nerf}
B.~Mildenhall, P.~P. Srinivasan, M.~Tancik, J.~T. Barron, R.~Ramamoorthi, and R.~Ng, ``Nerf: Representing scenes as neural radiance fields for view synthesis,'' \emph{Communications of the ACM}, vol.~65, no.~1, pp. 99--106, 2021.

\bibitem{sitzmann2019scene}
V.~Sitzmann, M.~Zollh{\"o}fer, and G.~Wetzstein, ``Scene representation networks: Continuous 3d-structure-aware neural scene representations,'' \emph{NeurIPS}, vol.~32, 2019.

\bibitem{atzmon2020sal}
M.~Atzmon and Y.~Lipman, ``Sal: Sign agnostic learning of shapes from raw data,'' in \emph{CVPR}, 2020, pp. 2565--2574.

\bibitem{chen2019learning}
Z.~Chen and H.~Zhang, ``Learning implicit fields for generative shape modeling,'' in \emph{CVPR}, 2019, pp. 5939--5948.

\bibitem{gropp2020implicit}
A.~Gropp, L.~Yariv, N.~Haim, M.~Atzmon, and Y.~Lipman, ``Implicit geometric regularization for learning shapes,'' \emph{arXiv preprint arXiv:2002.10099}, 2020.

\bibitem{li2022adaptive}
H.~Li, T.~Dai, Y.~Li, X.~Zou, and S.-T. Xia, ``Adaptive local implicit image function for arbitrary-scale super-resolution,'' in \emph{2022 IEEE International Conference on Image Processing (ICIP)}.\hskip 1em plus 0.5em minus 0.4em\relax IEEE, 2022, pp. 4033--4037.

\bibitem{chen2023cascaded}
H.-W. Chen, Y.-S. Xu, M.-F. Hong, Y.-M. Tsai, H.-K. Kuo, and C.-Y. Lee, ``Cascaded local implicit transformer for arbitrary-scale super-resolution,'' in \emph{CVPR}, 2023, pp. 18\,257--18\,267.

\bibitem{he2024latent}
Z.~He and Z.~Jin, ``Latent modulated function for computational optimal continuous image representation,'' in \emph{Proceedings of the IEEE/CVF Conference on Computer Vision and Pattern Recognition}, 2024, pp. 26\,026--26\,035.

\bibitem{shermeyer2019effects}
J.~Shermeyer and A.~Van~Etten, ``The effects of super-resolution on object detection performance in satellite imagery,'' in \emph{CVPR}, 2019, pp. 0--0.

\bibitem{wang2020dual}
L.~Wang, D.~Li, Y.~Zhu, L.~Tian, and Y.~Shan, ``Dual super-resolution learning for semantic segmentation,'' in \emph{CVPR}, 2020, pp. 3774--3783.

\bibitem{fookes2012evaluation}
C.~Fookes, F.~Lin, V.~Chandran, and S.~Sridharan, ``Evaluation of image resolution and super-resolution on face recognition performance,'' \emph{Journal of Visual Communication and Image Representation}, vol.~23, no.~1, pp. 75--93, 2012.

\bibitem{huang2024TSCUAP}
Y.~Huang, Q.~Guo, F.~Juefei-Xu, M.~Hu, X.~Jia, X.~Cao, G.~Pu, and Y.~Liu, ``Texture re-scalable universal adversarial perturbation,'' \emph{IEEE Transactions on Information Forensics and Security}, 2024.

\bibitem{huang2023ALA}
\BIBentryALTinterwordspacing
Y.~Huang, L.~Sun, Q.~Guo, F.~Juefei-Xu, J.~Zhu, J.~Feng, Y.~Liu, and G.~Pu, ``Ala: Naturalness-aware adversarial lightness attack,'' in \emph{Proceedings of the 31st ACM International Conference on Multimedia}, ser. MM '23.\hskip 1em plus 0.5em minus 0.4em\relax New York, NY, USA: Association for Computing Machinery, 2023, p. 2418–2426. [Online]. Available: \url{https://doi.org/10.1145/3581783.3611914}
\BIBentrySTDinterwordspacing

\bibitem{jia2020adv}
X.~Jia, X.~Wei, X.~Cao, and X.~Han, ``Adv-watermark: A novel watermark perturbation for adversarial examples,'' in \emph{Proceedings of the 28th ACM international conference on multimedia}, 2020, pp. 1579--1587.

\bibitem{zhang2020interpreting}
C.~Zhang, A.~Liu, X.~Liu, Y.~Xu, H.~Yu, Y.~Ma, and T.~Li, ``Interpreting and improving adversarial robustness of deep neural networks with neuron sensitivity,'' \emph{IEEE Transactions on Image Processing}, vol.~30, pp. 1291--1304, 2020.

\bibitem{li2023vrifle}
X.~Li, C.~Yan, X.~Lu, Z.~Zeng, X.~Ji, and W.~Xu, ``Inaudible adversarial perturbation: Manipulating the recognition of user speech in real time,'' in \emph{In the 31st Annual Network and Distributed System Security Symposium (NDSS)}, 2024.

\bibitem{li2023enroll}
X.~Li, J.~Ze, C.~Yan, Y.~Cheng, X.~Ji, and W.~Xu, ``Enrollment-stage backdoor attacks on speaker recognition systems via adversarial ultrasound,'' \emph{IEEE Internet of Things Journal}, vol.~11, no.~8, pp. 13\,108--13\,124, 2023.

\bibitem{biggio2013evasion}
B.~Biggio, I.~Corona, D.~Maiorca, B.~Nelson, N.~{\v{S}}rndi{\'c}, P.~Laskov, G.~Giacinto, and F.~Roli, ``Evasion attacks against machine learning at test time,'' in \emph{Machine Learning and Knowledge Discovery in Databases: European Conference, ECML PKDD 2013, Prague, Czech Republic, September 23-27, 2013, Proceedings, Part III 13}.\hskip 1em plus 0.5em minus 0.4em\relax Springer, 2013, pp. 387--402.

\bibitem{ze2023ultrabd}
J.~Ze, X.~Li, Y.~Cheng, X.~Ji, and W.~Xu, ``Ultrabd: Backdoor attack against automatic speaker verification systems via adversarial ultrasound,'' in \emph{2022 IEEE 28th International Conference on Parallel and Distributed Systems (ICPADS)}.\hskip 1em plus 0.5em minus 0.4em\relax IEEE, 2023, pp. 193--200.

\bibitem{xie2019improving}
C.~Xie, Z.~Zhang, Y.~Zhou, S.~Bai, J.~Wang, Z.~Ren, and A.~L. Yuille, ``Improving transferability of adversarial examples with input diversity,'' in \emph{CVPR}, 2019, pp. 2730--2739.

\bibitem{dong2019evading}
Y.~Dong, T.~Pang, H.~Su, and J.~Zhu, ``Evading defenses to transferable adversarial examples by translation-invariant attacks,'' in \emph{CVPR}, 2019, pp. 4312--4321.

\bibitem{lin2019nesterov}
J.~Lin, C.~Song, K.~He, L.~Wang, and J.~E. Hopcroft, ``Nesterov accelerated gradient and scale invariance for adversarial attacks,'' \emph{arXiv preprint arXiv:1908.06281}, 2019.

\bibitem{gao2020patch}
L.~Gao, Q.~Zhang, J.~Song, X.~Liu, and H.~T. Shen, ``Patch-wise attack for fooling deep neural network,'' in \emph{ECCV}.\hskip 1em plus 0.5em minus 0.4em\relax Springer, 2020, pp. 307--322.

\bibitem{long2022frequency}
Y.~Long, Q.~Zhang, B.~Zeng, L.~Gao, X.~Liu, J.~Zhang, and J.~Song, ``Frequency domain model augmentation for adversarial attack,'' in \emph{ECCV}.\hskip 1em plus 0.5em minus 0.4em\relax Springer, 2022, pp. 549--566.

\bibitem{zhou2018transferable}
W.~Zhou, X.~Hou, Y.~Chen, M.~Tang, X.~Huang, X.~Gan, and Y.~Yang, ``Transferable adversarial perturbations,'' in \emph{ECCV}, 2018, pp. 452--467.

\bibitem{huang2019enhancing}
Q.~Huang, I.~Katsman, H.~He, Z.~Gu, S.~Belongie, and S.-N. Lim, ``Enhancing adversarial example transferability with an intermediate level attack,'' in \emph{CVPR}, 2019, pp. 4733--4742.

\bibitem{zhang2022improving}
J.~Zhang, W.~Wu, J.-t. Huang, Y.~Huang, W.~Wang, Y.~Su, and M.~R. Lyu, ``Improving adversarial transferability via neuron attribution-based attacks,'' in \emph{CVPR}, 2022, pp. 14\,993--15\,002.

\bibitem{bevilacqua2012low}
M.~Bevilacqua, A.~Roumy, C.~Guillemot, and M.~L. Alberi-Morel, ``Low-complexity single-image super-resolution based on nonnegative neighbor embedding,'' 2012.

\bibitem{liang2022details}
J.~Liang, H.~Zeng, and L.~Zhang, ``Details or artifacts: A locally discriminative learning approach to realistic image super-resolution,'' in \emph{Proceedings of the IEEE/CVF Conference on Computer Vision and Pattern Recognition}, 2022, pp. 5657--5666.

\bibitem{xie2021learning}
W.~Xie, D.~Song, C.~Xu, C.~Xu, H.~Zhang, and Y.~Wang, ``Learning frequency-aware dynamic network for efficient super-resolution,'' in \emph{Proceedings of the IEEE/CVF International Conference on Computer Vision}, 2021, pp. 4308--4317.

\bibitem{zeyde2012single}
R.~Zeyde, M.~Elad, and M.~Protter, ``On single image scale-up using sparse-representations,'' in \emph{Curves and Surfaces: 7th International Conference, Avignon, France, June 24-30, 2010, Revised Selected Papers 7}.\hskip 1em plus 0.5em minus 0.4em\relax Springer, 2012, pp. 711--730.

\bibitem{martin2001database}
D.~Martin, C.~Fowlkes, D.~Tal, and J.~Malik, ``A database of human segmented natural images and its application to evaluating segmentation algorithms and measuring ecological statistics,'' in \emph{ICCV}, vol.~2.\hskip 1em plus 0.5em minus 0.4em\relax IEEE, 2001, pp. 416--423.

\bibitem{huang2015single}
J.-B. Huang, A.~Singh, and N.~Ahuja, ``Single image super-resolution from transformed self-exemplars,'' in \emph{CVPR}, 2015, pp. 5197--5206.

\bibitem{hore2010image}
A.~Hore and D.~Ziou, ``Image quality metrics: Psnr vs. ssim,'' in \emph{ICPR}.\hskip 1em plus 0.5em minus 0.4em\relax IEEE, 2010, pp. 2366--2369.

\bibitem{huang2011robust}
R.~Huang and K.~Sakurai, ``A robust and compression-combined digital image encryption method based on compressive sensing,'' in \emph{2011 Seventh international conference on intelligent information hiding and multimedia signal processing}.\hskip 1em plus 0.5em minus 0.4em\relax IEEE, 2011, pp. 105--108.

\bibitem{wikipedia_color_difference}
\BIBentryALTinterwordspacing
W.~contributors, ``Color difference,'' 2025, accessed: 2025-01-20. [Online]. Available: \url{https://en.wikipedia.org/wiki/Color_difference}
\BIBentrySTDinterwordspacing

\bibitem{wikipedia_jnd}
\BIBentryALTinterwordspacing
------, ``Just-noticeable difference,'' 2025, accessed: 2025-01-20. [Online]. Available: \url{https://en.wikipedia.org/wiki/Just-noticeable_difference}
\BIBentrySTDinterwordspacing

\bibitem{xie2020adversarial}
C.~Xie, M.~Tan, B.~Gong, J.~Wang, A.~L. Yuille, and Q.~V. Le, ``Adversarial examples improve image recognition,'' in \emph{CVPR}, 2020, pp. 819--828.

\bibitem{jia2024fast}
X.~Jia, J.~Li, J.~Gu, Y.~Bai, and X.~Cao, ``Fast propagation is better: Accelerating single-step adversarial training via sampling subnetworks,'' \emph{IEEE Transactions on Information Forensics and Security}, 2024.

\bibitem{li2023learning}
X.~Li, X.~Ji, C.~Yan, C.~Li, Y.~Li, Z.~Zhang, and W.~Xu, ``Learning normality is enough: a software-based mitigation against inaudible voice attacks,'' in \emph{32nd USENIX Security Symposium (USENIX Security 23)}, 2023, pp. 2455--2472.

\bibitem{li2024safeear}
X.~Li, K.~Li, Y.~Zheng, C.~Yan, X.~Ji, and W.~Xu, ``Safeear: Content privacy-preserving audio deepfake detection,'' in \emph{Proceedings of the 2024 on ACM SIGSAC Conference on Computer and Communications Security}, 2024, pp. 3585--3599.

\bibitem{xu2024uncovering}
T.~Xu, L.~Li, P.~Mi, X.~Zheng, F.~Chao, R.~Ji, Y.~Tian, and Q.~Shen, ``Uncovering the over-smoothing challenge in image super-resolution: Entropy-based quantification and contrastive optimization,'' \emph{IEEE Transactions on Pattern Analysis and Machine Intelligence}, 2024.

\bibitem{park2023perception}
S.~H. Park, Y.~S. Moon, and N.~I. Cho, ``Perception-oriented single image super-resolution using optimal objective estimation,'' in \emph{Proceedings of the IEEE/CVF Conference on Computer Vision and Pattern Recognition}, 2023, pp. 1725--1735.

\bibitem{qiu2023sc}
Z.~Qiu, Z.~He, Z.~Zhan, Z.~Pan, X.~Xian, and Z.~Jin, ``Sc-nafssr: Perceptual-oriented stereo image super-resolution using stereo consistency guided nafssr,'' in \emph{Proceedings of the IEEE/CVF Conference on Computer Vision and Pattern Recognition}, 2023, pp. 1426--1435.

\bibitem{zhang2018unreasonable}
R.~Zhang, P.~Isola, A.~A. Efros, E.~Shechtman, and O.~Wang, ``The unreasonable effectiveness of deep features as a perceptual metric,'' in \emph{Proceedings of the IEEE conference on computer vision and pattern recognition}, 2018, pp. 586--595.

\bibitem{wang2020scene}
W.~Wang, E.~Xie, X.~Liu, W.~Wang, D.~Liang, C.~Shen, and X.~Bai, ``Scene text image super-resolution in the wild,'' in \emph{ECCV}.\hskip 1em plus 0.5em minus 0.4em\relax Springer, 2020, pp. 650--666.

\bibitem{luo2019moran}
C.~Luo, L.~Jin, and Z.~Sun, ``Moran: A multi-object rectified attention network for scene text recognition,'' \emph{PR}, vol.~90, pp. 109--118, 2019.

\bibitem{deng2019retinaface}
J.~Deng, J.~Guo, Y.~Zhou, J.~Yu, I.~Kotsia, and S.~Zafeiriou, ``Retinaface: Single-stage dense face localisation in the wild,'' \emph{arXiv preprint arXiv:1905.00641}, 2019.

\bibitem{yang2016wider}
S.~Yang, P.~Luo, C.~C. Loy, and X.~Tang, ``Wider face: A face detection benchmark,'' in \emph{Proceedings of the IEEE conference on computer vision and pattern recognition (CVPR)}, 2016, pp. 5525--5533.

\bibitem{baidu_ocr}
\BIBentryALTinterwordspacing
Baidu, ``Baidu cloud ocr documentation,'' 2024. [Online]. Available: \url{https://cloud.baidu.com/doc/OCR/index.html}
\BIBentrySTDinterwordspacing

\bibitem{netzer2011reading}
Y.~Netzer, T.~Wang, A.~Coates, A.~Bissacco, B.~Wu, A.~Y. Ng \emph{et~al.}, ``Reading digits in natural images with unsupervised feature learning,'' in \emph{NIPS workshop on deep learning and unsupervised feature learning}, vol. 2011, no.~2.\hskip 1em plus 0.5em minus 0.4em\relax Granada, 2011, p.~4.

\end{thebibliography}


\end{document}